\documentclass[final,5p, lmodern]{elsarticle}

\usepackage[hidelinks]{hyperref}
\usepackage{xcolor}
\usepackage{soul}\setuldepth{article}
\usepackage{graphicx}
\usepackage{booktabs}
\usepackage{multirow}
\usepackage{multicol}
\usepackage{tabularx}
\usepackage{hyperref}    
\usepackage{amsfonts}    
\usepackage{amsmath}
\usepackage{amssymb}
\usepackage{caption}
\usepackage{subcaption}
\usepackage{rotating}
\usepackage{pdflscape}
\usepackage{bbm}
\usepackage{lineno}
\usepackage{nccmath}
\usepackage{placeins}
\usepackage{microtype}
\usepackage{orcidlink}

\journal{Artificial Intelligence in Medicine}

\begin{document}

\begin{frontmatter}



\title{Probabilistic emotion and sentiment modelling of patient-reported experiences}


\author[UoA]{Curtis Murray\orcidlink{0000-0001-6104-4785}\corref{cor1}}
\cortext[cor1]{Corresponding Author:}

\ead{curtis.murray@adelaide.edu.au}

\author[UoA]{Lewis Mitchell\orcidlink{0000-0001-8191-1997}}
\author[UoA]{Jonathan Tuke\orcidlink{0000-0002-1688-8951}}
\author[JC]{Mark Mackay\orcidlink{0000-0002-7759-8755}}

\affiliation[UoA]{
  organization={The University of Adelaide, School of Mathematical Sciences},
  addressline={North Terrace}, 
  city={Adelaide},
  postcode={5005}, 
  state={SA},
  country={Australia}
}

\affiliation[JC]{
  organization={James Cook University},
  addressline={1 James Cook Dr}, 
  city={Townsville}, 
  postcode={4811}, 
  state={Queensland},
  country={Australia}
}

\begin{abstract}

    Patient feedback is necessary to assess the extent to which healthcare delivery aligns with the needs and expectations of the public. Despite the utility of surveys in soliciting structured patient feedback that is readily analysed, they can be costly, infrequent, and their predefined scope inhibits a comprehensive understanding of patient experience. In contrast, online review websites and social media platforms' freedom from this structure may afford unique insights into patient perspectives, yet this lack of structure presents a challenge for analysis. In this study, we produce a novel methodology for interpretable probabilistic modelling of patient emotions from patient-reported experiences. We employed metadata network topic modelling to uncover and explore key themes in patient-reported experiences sourced from the popular healthcare review website Care Opinion. Insightful relationships between these themes and labelled emotions found in the text are revealed. Positivity and negativity are most strongly related to aspects of patient experience, such as patient-caregiver interactions, rather than outcomes of clinical care. Patient engagement through educational programs also exhibits strong positivity, whereas dismissal and rejection are linked with suicidality and depression. We develop a probabilistic, context-specific emotion recommender system capable of predicting both multilabel emotions and binary sentiments using a naive Bayes classifier using contextually meaningful topics as predictors. The superior performance of our predicted emotions under this model compared to baseline models was assessed using the information retrieval metrics nDCG and Q-measure, and our predicted sentiments achieved an F1 score of 0.921, significantly outperforming standard sentiment lexicons. This methodology provides a cost-effective, timely, transparent, and interpretable approach to harness and capture unconstrained patient-reported feedback, with the potential to augment traditional patient-reported experience collection methodologies. Our probabilistic emotion and sentiment recommender system is available in an R package and interactive dashboard for future research and clinical practice applications, making it a valuable tool for healthcare researchers looking to better understand and improve patient experiences. In particular, this approach can offer distinct and immediate value when incorporated in surveys with free-text comments to capture a patient's emotional well-being. By gaining a complementary understanding of patient emotions and needs, hospitals can tailor individualised care to more effectively meet the public's expectations. Overall, our study presents a novel approach to harness the power of unconstrained patient-reported feedback, providing healthcare researchers with valuable insights that can ultimately improve patient experiences and enhance the quality of healthcare delivery.
\end{abstract}




\end{frontmatter}


\section{Introduction}

\subsection{Background on Patient-Reported Experience}

Patient-reported outcomes (PROs) and patient-reported experiences (PREs) are increasingly being used to develop a more holistic view of healthcare in a patient-centred care approach \cite{commonwealthofaustralia_2019}. PROs and PREs can provide valuable insights into the quality and effectiveness of healthcare services and their alignment with patient expectations \cite{nemeth2006health}. As such, patient-reported outcome measures (PROMs) and patient-reported experience measures (PREMs) are being adopted as important indicators of healthcare performance \cite{prosandprems, NHIPCC2017}. 

\subsection{Traditional Acquisition Methods and their Limitations}

Despite their increasingly recognised importance in medical literature, patient-reported experience acquisition practices, such as surveying and focus groups, may fail to capture a complete picture of patient-reported experience. The power of a survey lies in its ability to methodically gather answers to specific questions posed to its audience. This is achieved through the structured format that lends itself to rapid analysis of these specific questions. The specificity and structure of a survey, while appealing, inhibits the exploration and understanding of issues that are not explicitly addressed in the survey \cite{simon2013scope}. Questions that are not posed in a survey have no way to be answered, and survey-style multiple-choice questions may fail to capture the complexity of the patient experience. In this way, the most common technique for collecting patient-reported experiences suppresses the patient from voicing their full experience. On the other end of the spectrum are focus groups. This qualitative-driven approach to harnessing opinions can succeed in providing scope to discuss issues the participants are interested in \cite{fitzpatrick1996qualitative}. Where this approach suffers, however, is in its lack of scalability \cite{vicsek2010issues}, and personal nature that may lead to self-censorship \cite{fitzpatrick1996qualitative}. This leads to focus groups often being used as qualitative explorations used to drive survey questions \cite{morgan1996focus}. There is a need to extend how we currently harness patient-reported experiences, to have enough relevance to capture issues that the patient wants to voice, in a cheap, scalable way. 

\subsection{Emergence of Online Patient Feedback and Challenges}

Recently, increasing effort has been directed online, to \textit{the cloud of patient experience} as a means to provide real-time and low-cost methods to explore healthcare \cite{Greaves2013}. Online, uninhibited expositions document patient experiences without the constraints imposed by surveys, providing richer representations of experiences. However, the lack of structure in these expositions presents challenges in large-scale analysis.

\subsection{Online Patient Feedback Analytical Methods}
The lack of structure in free-text means that conventional surveying techniques do not apply. Instead, researchers turn to Natural Language Processing (NLP), to elicit latent structure from the reports \cite{births, okon2020natural, du2020leaked,murray2021symptom, hawkins2016measuring, bovonratwet2021natural, clark2018sentiment}. This latent structure, once uncovered can be exploited to capture patient-reported experiences for individuals, as well as varying strata. Sentiment analysis is a popular NLP technique that attributes sentiments to documents. Approaches in sentiment analysis typically use a sentiment lexicon, a vocabulary that associates sentiments with words to summarise a document's sentiment. Alternatively, machine learning approaches can be used to train a sentiment analysis model. Without labelled training data, pre-trained models known as language transformers are often used. While the simplicity of sentiment lexicons facilitates their easy adoption, their over-generalised nature inhibits their ability to accurately model sentiments in varied contexts. Large language models (LLMs), such as BERT \cite{devlin2018bert}, and OpenAI's GPT models \cite{radford2018improving,radford2019language, brown2020language, ouyang2022training} are often considered as \textit{black-boxes} that lack interpretability, and may also suffer from a lack of contextual awareness if not fine-tuned. Moreover, these models suffer from intrinsic biases from the data they are trained on, and often \textit{hallucinate} untrue responses \cite{wake2023bias, alkaissi2023artificial}. These issues collectively contribute to a mistrust in the application of LLMs for healthcare applications \cite{wang2023ethical}.

Ensuring the interpretability of a model fosters trust in its results. This is crucial for the successful implementation and utilisation of models in healthcare, where the delivery of healthcare services has high potential to significantly impact health and well-being. As such, scalable patient-reported experience mining must be met with interpretability to foster trust that leads to its successful adoption in patient-reported experience measures. 

\subsection{Study Objectives and Findings}

The website Care Opinion \cite{careopaus} is an independent not-for-profit charitable institution that publishes user-submitted reports of patient experiences related to Australian hospitals. These reports contain both feedback on the healthcare system in Australia in free-text comments, as well as labelled data on aspects of patient experience, such as emotions.

This study aims to develop and evaluate a methodology for capturing and harnessing unconstrained patient-reported feedback in healthcare, in a manner that is cost-effective, timely, transparent, interpretable, and accurate. In this paper, we use metadata network topic modelling \cite{hyland2021multilayer} as a dimension reduction to the Care Opinion data to uncover latent structure in patient-reported experiences. We reveal relationships between themes of discussion and emotions by finding topic profiles for each emotion. We further capture relationships between topics and sentiments (positive and negative) through topic-sentiment profiles, where sentiments are manually classified from emotions. This enables us to quantify the positivity and negativity of each topic in the discourse. We explore the most positive, and most negative topics in the discourse and reveal that positivity and negativity are most strongly related to aspects of patient experience, such as interactions with healthcare workers, rather than outcomes of clinical care. We show the relationships between these most positive and most negative topics and the collection of labelled emotions. We find the most negative topic is most associated with the labelled emotions \textit{abused} and \textit{ridiculed}, and the most positive topic most associated with \textit{admiration} and \textit{amazed}. We reveal a landscape of emotions through a dimension reduction of emotion-topic representations, that visualises the similarity of emotions through their proximity. Clusters of positive and negative emotions emerge that agree with our manual emotion-sentiment labelling. We contribute to the research problem of capturing patient-reported experience in a cheap, scalable, and interpretable manner by developing methodology and implementation of a context-specific probabilistic emotion recommender system. This is done through an application of the naive Bayes algorithm, using the topic modelling dimension reduction of the care opinion corpus as features in the prediction of the labelled emotion responses. We show that using this dimension reduction, as opposed to an approach considering the full text, prevents over-fitting, and yields superior results for both multi-label emotion prediction and binary sentiment prediction under appropriate metrics through 10-fold cross-validation. Using these topics as contextually meaningful predictors in a probabilistic Naive Bayes framework affords a high degree of interpretability and ensures model predictions are transparent. Further, we show that this model exhibits high performance at binary sentiment classification, outperforming standard sentiment lexicons. We anticipate the application of our model into healthcare through its incorporation into an R package \cite{persR} and online dashboard \cite{persShiny}, both of which may easily be adopted by healthcare researchers and practitioners seeking to uncover latent emotions in free-text patient reports to better capture patient-experience.

\section{Methods}

\subsection{Care Opinion Data}
A corpus of 13,380 reports from February 2012 to October 2022 containing patient-reported experiences in hospitals was collected from the popular review website Care Opinion (formerly known as Patient Opinion). Each report in the Care Opinion corpus contains information on the title, the report text, the date reported, the location of the report, tags, and prompted answers to \textit{What's good?}, \textit{What could be improved?}, and \textit{Feelings}.

While there is structure in aspects of the Care Opinion reports, such as a catalogue of prompted answers to questions, the free text comments about healthcare contain detailed depictions of patient experience that may be otherwise overlooked. Relationships between free-text comments and labelled responses to prompted questions offer a semi-structured source for an understanding of patient-reported experiences in healthcare through natural language processing. Natural language processing tools such as topic modelling and sentiment analysis provide methods to uncover latent structure in free-text comments. This structure can provide an overview of what is being discussed, and particular sentiments conveyed.

\subsection{Methods to Analyse Patient Narratives on Care Opinion}

In the analysis of Care Opinion reports, we employ the Design-Acquire-Process-Model-Analyse-Visualise (DAPMAV) framework, which we previously introduced, to systematically approach the natural language processing of patient narratives \cite{murray2023revealing}. The framework guides us through designing the study around specific patient experience themes, processing the text to identify meaningful patterns, modelling to uncover latent structures, analysing the relationships between topics and sentiments, and visualising the results for actionable insights.

\subsubsection{Proprocessing}

In the preprocessing stage, we normalised the patient narratives to lowercase, contracted hyphenated words and conjunctions. All remaining non-alphabetic characters were replaced with spaces. Additionally, we removed words occurring fewer than five times to reduce noise and enhance data quality for our analysis. Further, we remove stop words (words with little-to-no contextual meaning such as \textit{the}, \textit{it}, and \textit{and}), from the R \textit{tidytext} package \cite{silge2017text}. Narratives where no patient-reported emotions were present were removed from the corpus prior to topic modelling, resulting in $10,509$ patient-reported experience narratives.

\subsubsection{Topic Modelling}

Different themes of a patient's experience in healthcare may be reported online. In topic modelling, we attempt to model themes in text through collections of words that appear in statistically similar ways, called \textit{topics}. The objective of a topic model is to express documents, or in this case, patient reports, as mixtures of meaningful topics that capture a specific theme of conversation, where the topics themselves are mixtures of words. This expression of documents as mixtures of topics acts as a dimension reduction, where documents, originally embedded in the high-dimensional ($\approx10,000$) vocabulary space known as a bag-of-words, are mapped to a relatively low-dimensional ($\approx 100$) topic-space. By summarising the documents in this manner, we obtain an insightful representation of the data that would otherwise be unnecessarily specific and hence cumbersome for a general overview.

\begin{figure*}[!h]
    \centering
    \includegraphics[width=1\textwidth]{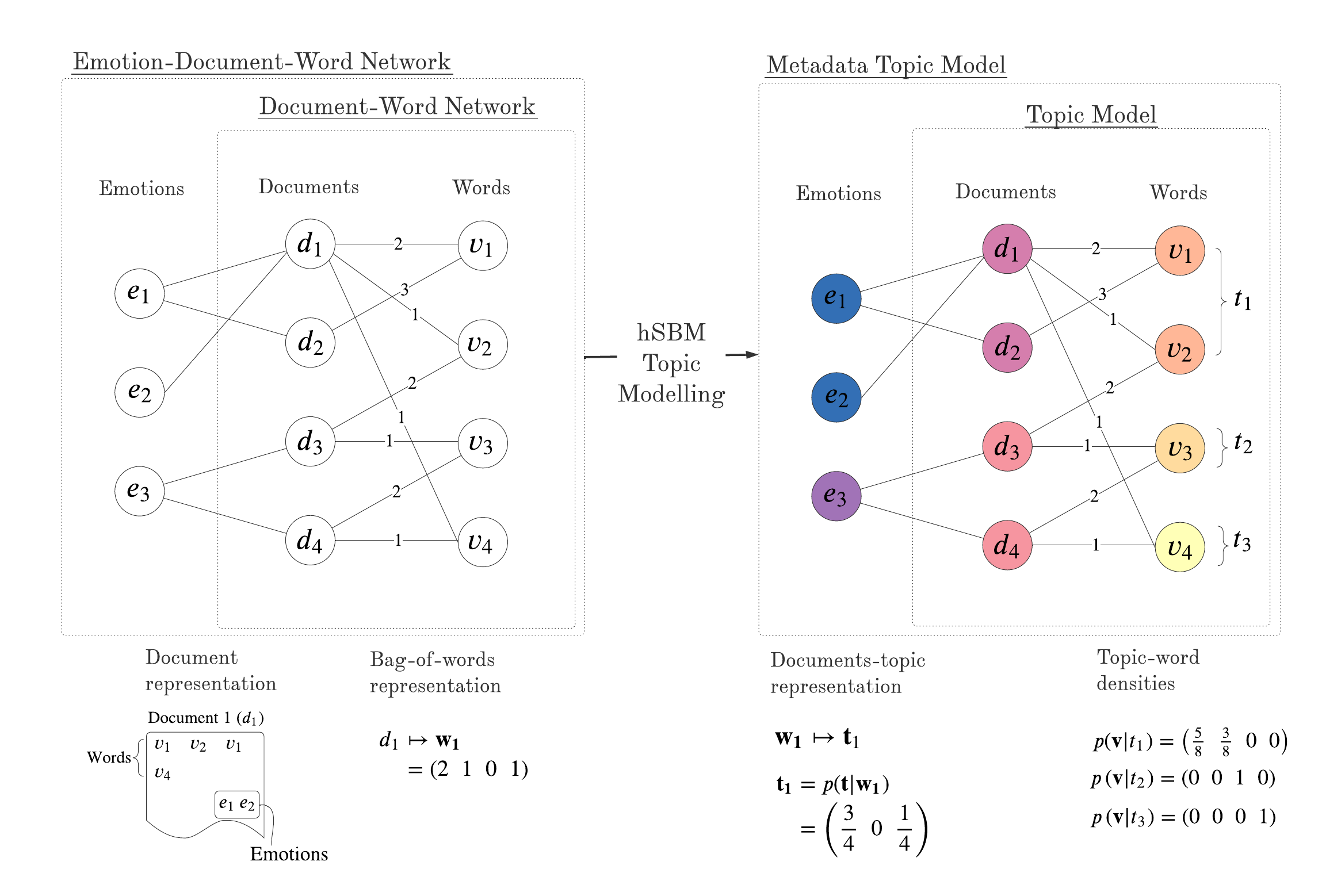}
    \caption{\textbf{Metadata Topic Modelling Overview:} Documents are represented in a document-word network where edges between document nodes and word nodes count the number of occurrences of word $v_i$ in document $d_j$ (left). Emotion labels $e_t$ are added to this network in the emotion-document-word network by connecting document nodes to emotion nodes (with an implicit edge count of one). Below this network Document $1$ is depicted as consecutive words $v_1,v_2, v_1, v_4$, and tagged with emotions $e_1, e_2$. We also show the bag-of-words representation as the vector of word counts. Both networks undergo hSBM topic modelling. The hSBM performs community detection, which we visualise through the colouring of nodes to indicate group membership. Topics are the communities of word nodes. We show the mapping from $d_1$, or its bag of words representation $\mathbf{w_1}$ to the document-topic representation $\mathbf{t_1} = p(\mathbf{t} | \mathbf{w_1})$, which is taken as the empirical densities of topic use. To find this, we calculate the denominator as the number of edges out of the document $(4)$, and numerators as the number of edges from the document to each respective topic $(3, 0, 1)$. The corresponding topic-word densities that indicate $p(\mathbf{v} | t_i)$ are again the empirical densities, taken as the number of edges to the topics $(8,1,1)$ as denominators, and the number of edges to each word as the numerators ($(5,3,0,0)$, $(0,0,1,0)$, and $(0,0,0,1)$) for topics $t_1$, $t_2$, and $t_3$ consecutively. In \ref{appendix:example} we show the calculation of the posterior distribution $p(E=e | d_1)$.\label{fig:model-example}}
\end{figure*}

Recent developments in topic modelling \cite{gerlach2018network} show that network topic modelling generalises the probabilistic Latent Semantic Indexing (pLSI) objective \cite{hofmann1999probabilistic} of traditional latent Dirichlet allocation (LDA) topic models \cite{blei2003latent} to allow for hierarchical topic structures, while avoiding the unjustified uni-modal Dirichlet distribution that is not representative of real text; provide a Bayesian approach to choose the number of topics, and, from the lens of minimum description length, provides a more concise compression of the data. Network topic modelling functions by performing community detection with hierarchical stochastic block models (hSBMs) on a document-word network \cite{gerlach2018network} to cluster words into topics. Network topic modelling using hSBMs can be extended to consider additional metadata in documents with the addition of new nodes corresponding to the metadata that links to documents \cite{hyland2021multilayer}. For example, patient-reported emotion metadata can be linked to documents by using a metadata topic model, allowing this information to be pooled into the topic model. Figure \ref{fig:model-example} conceptually summarises network topic modelling, and metadata network topic modelling, showing a clustering of the document-word network (with patint-reported emotion metadata).

\subsubsection{Relationships between Topics and Patient-Reported Emotions}

In our analysis, we quantitatively link patient-reported emotions to the thematic structures extracted from healthcare narratives using topic modelling. This is achieved by marginalising document-topic distributions, denoted as 
$p(T=t | d)$, across all documents tagged with a specific emotion. Here we use $T$ to represent a random topic variable, $k$ a specific topic, and $d$ a specific document. The emotion-topic density for each topic $k$ given an emotion $e$ is computed as

\begin{align}
    \label{eq:topic-emotion-density}
p(T = k | E = e) = \frac{1}{|D_e|} \sum_{d \in D_e} p(T=k | d)
\end{align}
where $|D_e|$ is the count of documents associated with emotion $e$. This approach enables us to systematically associate each topic with corresponding emotional responses, providing a nuanced understanding of patient experiences in healthcare settings.

The likelihood of a topic given \textit{any} positive (or negative) emotion is found by marginalising the joint emotion-topic density over the positive (or negative) emotions,
\begin{align}
    \label{eq:topic-sentiment-density}
    p(T = k | E\in E_p) &= \sum_{e_i \in E_p}p(T= k | E = e_i)p(E = e_i | E \in E_p) \nonumber \\
    &= \dfrac{\sum_{e_i \in E_p}p(T= k | E = e_i)p(E = e_i)}{\sum_{e_i \in E_p}P(E = e_i)}.
\end{align}

We define the \textbf{positivity} of a topic as the ratio of positive topic likelihoods to negative topic likelihoods,
\begin{align}\label{eq:ratio}
  \mathrm{Positivity}(k) =  \frac{p(T=k |E\in E_p)}{p(T=k |E\in E_n)},
\end{align}
which tells us how many times more likely topics are to be used with positive emotions than with negative ones. Topic \textbf{negativity} is defined similarly, being the inverse of topic positivity. Topics that have a high positive-to-negative topic ratio, or positivity, are associated more often with positive sentiments, and topics with a low positive-to-negative topic ratio, conversely, are associated more often with negative sentiments. 

\subsubsection{A Landscape of Patient-Reported Emotions}

The emotion-topic profiles provide summaries of the thematic compositions of experience narratives associated with each emotion. These thematic compositions, when shared across different emotions, suggest a commonality in the underlying patient experiences. Uncovering relationships between emotions that are reflective of common experiences can help to reveal collective patient experience in more detail. This is especially true when considering that some emotions, such as \textit{happy} and \textit{sad} describe one's intrinsic experience, whereas emotions such as \textit{rejected} or \textit{empowered} describe an extrinsic experience that shapes one's feelings. Recognising relationships between experiences that result in intrinsic and extrinsic emotions allows a deeper comprehension of intrinsic emotions in the context of patient experience. To investigate the thematic relationships, we examine the distances between emotion-topic densities. Applying Uniform Manifold Approximation and Projection (UMAP) \cite{mcinnes2018umap} to these densities, we represent the emotion-topic space in a two-dimensional projection that preserves local distance. Consequently, emotions with thematic similarities are positioned in proximity within this projection.

\subsection{Modelling Patient Emotions from Patient Experience Narratives}

\subsubsection{Sentiment Analysis}

The natural language processing tool \textit{sentiment analysis} allows for the detection and quantification of a text's \textit{sentiment} by either inferring the text's positivity or identifying the emotions it presents. A simple methodology to do so involves comparing a body of text to a sentiment lexicon, such as AFINN, which expresses the sentiment of words as integer values ranging from -4 (most negative) to 4 (most positive) \cite{NielsenF2011New}. Other popular sentiment lexicons include VADER \cite{hutto2014vader}, SentiWordNet \cite{baccianella-etal-2010-sentiwordnet}, NRC \cite{mohammad2013crowdsourcing}, and Bing \cite{hu2004mining}. Dictionary approaches such as this are easily implemented, and can be effective; however, general dictionary approaches often fail to account for contextual information that may alter the sentiment of the words being used.  In medical reports, context can significantly alter the implied sentiment of words and phrases. For instance, a \textit{positive} diagnosis is paradoxically not positive, suggesting the presence of a condition or abnormality. Similarly, the word \textit{intense} can convey contrasting sentiments based on context: in a review of a workout program, \textit{intense} might be perceived positively, reflecting a high level of challenge and effectiveness, whereas, in patient feedback about pain experience, \textit{intense} is indicative of severe discomfort, a decidedly negative sentiment. The contrast in sentiments conveyed by identical words across different contexts underscores the challenges and complexities inherent in sentiment analysis.

Recognising that sentiment varies through context is a challenge for sentiment analysis. The website Care Opinion provides prompted answers for patients to express their emotions. Associating these labelled emotions with the respective free-text comments provides the possibility for the development of emotion and sentiment analysis models for detecting sentiments from patient-reported experiences. We develop one such model in the following section.

\subsubsection{Probabilistic Emotion Recommender System}\label{section:model}

In this paper we construct a context-specific probabilistic recommender system for emotions from the viewpoint of patients and their experiences in hospitals from the Care Opinion reports by exploiting the relationship between free-text reports and labelled responses to prompted questions. We find estimates for the posterior distribution of emotions given a new patient-reported experience narrative bag-of-words,
\begin{equation}
    p(E = e  | \mathbf{w})
\end{equation}
through a naive Bayes approach, assuming class-conditional independence of words. We use topic modelling as a dimension reduction tool to (1) extract contextually meaningful predictors that (2) reduce the high dimensionality of the text data, associating reduced topic spaces with emotions. This feature engineering approach aims to capture more general relationships between patient experiences and emotions by pooling information from contextually similar words into themes. By doing so, overfitting from sparse interactions between emotions and words is mitigated, and meaningful interpretation of how model predictions are made is fostered. To ensure numerical stability in our calculations, techniques like the log-sum-exp trick are utilised. The model also incorporates practical adjustments to maintain stability and accuracy. For a detailed exposition of the model's mathematical framework, readers are referred to Appendix \ref{app:pers}.

This methodology provides a statistical foundation for a context-specific, probabilistic emotional recommender system, where recommendations are given as the top-ranking emotions under the posterior distribution.

\subsubsection{Binary Sentiment Classification of Patient Reports}

We can further benefit from this model by using it for probabilistic sentiment analysis where we only consider how positive (or negative) documents are by marginalising the posterior over positive (or negative) emotions.

\begin{align}\label{eq:binary}
    p(E \in E_p | \mathbf{w}) = \sum_{e_i \in E_p} {p(E = e_i | \mathbf{w})},
\end{align}
where $E_p$ is the set of positively labelled emotions (and $E_n$ the set of negatively labelled emotions).

Since $E_p$ and $E_n$ partition the set of emotions, 
$$
P(E \in E_n | \mathbf{w})= 1 - P(E \in E_p | \mathbf{w}). 
$$

We perform binary classification of a document's sentiment as either positive or negative by hard classifying to the highest density sentiment class according to the marginalised posterior of Equation \ref{eq:binary},
\begin{equation}
    \label{eq:hard}
    \hat{S} = \displaystyle \operatorname*{argmax}_{c\in \{p,n\}} P(E\in E_c | \mathbf{w}).
\end{equation}
As there are patient narratives labelled with both positive and negative emotions, we hard-classify similarly to Equation \ref{eq:hard} using the empirical data. In both instances, we deal with ties by classifying a post as positive, reflecting the overall more prevalent class in the Care Opinion corpus.

\subsubsection{Model Evaluation}

We employed a $k$-fold cross-validation method, partitioning the Care Opinion data into ten equal folds. Within each fold, we trained three iterations of the probabilistic recommender system, as described in Section \ref{section:model}. Two of these iterations utilise hierarchical Stochastic Block Models (hSBM) — a standard topic model and a metadata-enhanced variant — to cluster words into topics. The third iteration, in contrast, bypasses this dimension reduction step and uses the full vocabulary, as per the model detailed in Equation \ref{eq:word}. 

To establish baseline comparisons, two additional models were incorporated. The first is a Maximum-Likelihood Estimate (MLE) model, which assigns emotion probabilities based on the empirical densities of emotion classes observed in the training set. This model serves as a baseline that takes into account the imbalances in emotion classes. The second baseline model adopts a uniform-random approach, assigning equal probability to all emotion classes, thereby not compensating for class imbalances.

We evaluate the performance of the recommender systems using precision and recall, interpolated precision at $k$, as well as Q-measure and normalised Discounted Cumulative Gain (nDCG). In all cases, we report the mean metric across all $10$ folds. Both Q-measure and nDCG incorporate partial relevance of misclassifications into their model evaluations. This is particularly beneficial as there are over 200 patient-reported emotions present in the Care Opinion corpus. Considering partial relevance in model evaluations allows for lower penalisation when partially correct emotions are predicted, as opposed to when completely irrelevant emotions are predicted, through a  \textbf{relevance} metric.  We define the relevance, or equivalently gain, of a predicted emotion $e_p$ given a true labelled emotion $e$ as
\begin{align}\label{eq:rel}
\mathrm{rel}(e_p, e) = \begin{cases} \frac{\max_i(d(e_i, e)) - d(e_p, e)}{\max_i(d(e_i, e))} &\text{ if }C(e_p) = C(e) \\
0 &\text{ otherwise},
\end{cases}
\end{align}

where $C(e)$ is an indicator of the positivity of $e$, 
\begin{align*}
C(e) = \begin{cases} 1 \text{ if $e$ is positive}  \\
0 \text{ otherwise},
\end{cases}
\end{align*}
and $d(\cdot,\cdot)$ is the Euclidean distance between topic-emotion densities,
\begin{align*}
    d(e, e_p) &= \| p(\mathbf{t} | e) - p(\mathbf{t} | e_p) \| \\
    &= \sqrt{\sum_{k = 1:n_t} \left(p(T=k | e) - p(T=k | e_p)\right)^2}. 
\end{align*}

The relevance score \( \mathrm{rel}(e_p, e) \) quantifies the closeness of a predicted emotion to the true emotion by normalising the emotion distance to a $0$-to-$1$ scale, assigning a relevance of $1$ for an exact match and $0$ for the greatest emotion distance, while also truncating the score to 0 for any predicted emotions that differ in sentiment classification from the true emotion, reflecting zero relevance. Full definitions for Q-measure and nDCG are deferred to \ref{app:eval}. 

The models we consider are adapted into binary sentiment classifiers as outlined previously, and their performances are evaluated using accuracy, balanced accuracy, and the macro-averages of the F1 score, precision, and recall. Additionally, we benchmark against other sentiment analysis models --- namely, Bing, NRC, SentiWordNet, VADER, and AFINN --- by calculating the average sentiment score per word and assigning the document to the most dominant sentiment class. In cases where positive and negative sentiments have equal weight, we default to classifying the document as positive, the most prevalent sentiment class in the Care Opinion corpus.

\section{Results}

\subsection{Analysis of Patient Narratives on Care Opinion}

\begin{figure}[h]
    \centering
    \centering
    \includegraphics[width=0.4\textwidth]{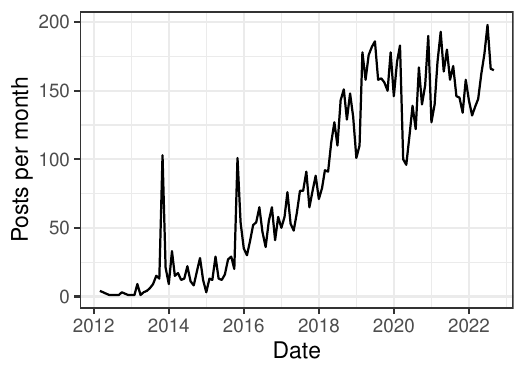}
    \captionof{figure}{Number of monthly reports to Care Opinion.}
    \label{fig:monthly_posts}
    \end{figure}

Figure \ref{fig:monthly_posts} illustrates the rise in popularity of reports to Care Opinion, growing from approximately 2 reports per month in 2012 to approximately 160 reports per month in 2019. We see a sharp drop in monthly reports at the start of 2020 that corresponds with the COVID-19 pandemic. The COVID-19 pandemic had a significant impact on access to some healthcare services in Australia due to restrictions on travel, in combination with strains on other aspects of healthcare from increased demand for COVID-19 testing and treatment \cite{parliament_australia_2022_covid}.


    \begin{figure}[h]
    \centering
    \includegraphics[width=0.4\textwidth]{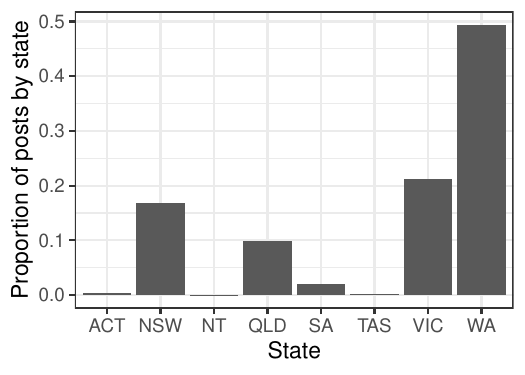}
    \captionof{figure}{Number of reports to Care Opinion by Australian states and territories}
    \label{fig:states}
   \end{figure}

   In Figure \ref{fig:states}, we observe a noticeable disparity in the distribution of reports across Australian states, with a notably higher proportion originating from Western Australia. This trend suggests a potentially greater awareness or adoption of Care Opinion in this region among healthcare practitioners and the public. Such regional variations in engagement with the platform can provide insights into differing levels of digital health literacy and public health communication strategies across states.
   
   The prompted answers to the \textit{Feelings} tag, for example, \textit{happy}, \textit{disappointed}, and \textit{thankful}, associate emotions with patient experience narratives. We see the use of $263$ distinct emotions in the Care Opinion corpus, which we manually classify as either positive or negative. In total there were $26163$ tags made, $58.7\%$ of which were positive, and the remaining $41.3\%$ negative. Posts tagged with solely positive sentiments make up $43.4\%$ of posts. Solely negatively tagged posts make up $23.4\%$ of the corpus. Mixed posts make up $5.4\%$ of the corpus, with average positivity, i.e. average proportion of tags that are positive being $0.539$. There are also an additional $21\%$ of posts that do not contain tags. This is summarised in Table \ref{tab:classification}. These results correspond to those found in those found for a smaller study of $427$ reports in the UK version of Care Opinion \cite{ramsey2019healthcare, careopuk}. There, through manual sentiment analysis, they found $59.7\%$ positively classified reports, $23.7\%$ negatively classified reports, and $16.7\%$ mixed reports. 
   
   
   
   \begin{table}[h]
    \centering
    \caption{Summary of positive, negative, and mixed posts. Positive/negative posts have entirely positive/negative tagged emotions, whereas mixed is a combination. The positivity of a post is defined here as the proportion of tagged emotions that are positive, with the mean of each type of post reported. The total number (count) of each type of post is given, as well as the proportion of each type of post.}
       \label{tab:classification}
   \footnotesize
   \begin{tabular}{>{\raggedright\arraybackslash}p{4em}cccc}
   \toprule
   Sentiment & Positivity & Count & Proportion & Tags per post\\
   \midrule
   Positive & 1.000 & 5807 & 0.434 & 2.355\\
   Negative & 0.000 & 3993 & 0.234 & 2.358\\
   Mixed & 0.539 & 719 & 0.054 & 4.267\\
   None &  - & 2861 & 0.210 & 0\\
   \bottomrule
   \end{tabular}
   \end{table}
   
   While there is an overall tendency for posts to be positive, and the mean number of tags per post is comparable for both wholly positive and negative posts, we observe a more expansive vocabulary of tagged emotions for negative sentiment posts (96 unique positive emotions and 167 unique negative emotions); there are nearly twice the number of unique negative sentiments than positive sentiments. This phenomenon is not unique to this corpus, nor even the English language; a paper that analysed multiple sources in three languages noticed that words with negative emotions are used less than words with positive emotions, but because of their rareness they carry more information \cite{garcia2012positive}.
   
   \begin{figure}[h]
       \centering
        \includegraphics[width=.4\textwidth]{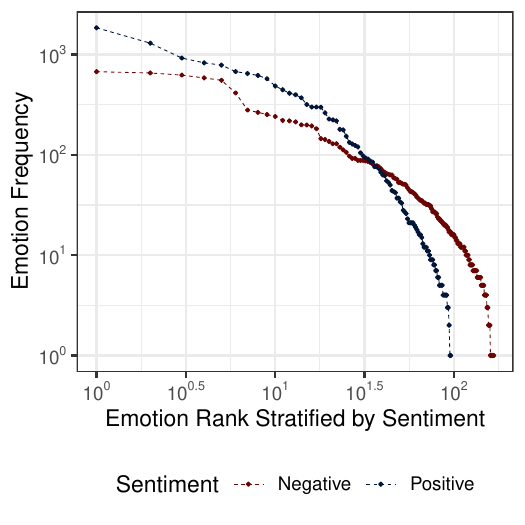}
        \captionof{figure}{Log-Log Distribution of Emotion Frequencies Stratified by Sentiment in Patient Feedback. }
        \label{fig:loglogrank}
       \end{figure}
   
   Figure \ref{fig:loglogrank} shows a significantly heavier tail in the rank-size distribution for negative sentiments than in the positive counterpart. This indicates that, overall, a smaller proportion of people tend to make negative posts, although those who do tend to express themselves using a richer vocabulary through their tagged emotions.
   
\subsubsection{Topic Themes in Care Opinion Reports}

Our network topic modelling analysis using a hierarchical stochastic block model on the Care Opinion patient reports reveals 105 distinct topics at the deepest level in the topic hierarchy. In Table \ref{tab:topics} we show the ten most prevalent topics with the five most common words in each, along with their respective topic densities across the corpus. In general, we observe that topics broadly fall under the following three themes, with some topics relating to more than one general theme at once.

\begin{table}[h!]
    \centering
    \caption{Most prevalent topics and their frequencies in patient narratives from Care Opinion}
    \label{tab:topics}
    \footnotesize
\begin{tabular}{>{\raggedright\arraybackslash}p{20em}>{\centering\arraybackslash}p{4em}>{}p{4em}>{}p{4em}>{}p{4em}}
\toprule
Topic & Density\\
\midrule
information, system, late, required, concerned & 0.041\\
found, im, lot, issues, hope & 0.040\\
extremely, level, spent, difficult, understanding & 0.037\\

wonderful, professionalism, kindness, appreciated, exceptional & 0.035\\
amazing, grateful, comfortable, fantastic, safe & 0.034\\

time, day, times, feeling, informed & 0.033\\

opinion, dont, understand, leave, speak & 0.030\\
didnt, couldnt, wasnt, finally, happened & 0.028\\
caring, professional, friendly, excellent, impressed & 0.027\\

night, morning, arrived, explained, busy & 0.025\\
service, team, happy, helpful, provided & 0.025\\
support, helped, recovery, knowledge, journey & 0.020\\

upset, recall, disappointed, telling, requested & 0.020\\
hospital, doctors, hospitals, regional, performed & 0.020\\
due, offered, completely, allowed, previous & 0.019\\

health, services, centre, community, provide & 0.018\\

blood, symptoms, breathing, cold, bleeding & 0.018\\
staff, nursing, pass, cleaning, polite & 0.018\\

recently, received, attended, short, attention & 0.017\\
private, lack, request, form, mentioned & 0.016\\
\bottomrule
\end{tabular}
\end{table}

\textbf{Clinical Care, Procedures, Recovery, Rehabilitation, and Outcomes}: This theme of topics encompasses interactions with healthcare professionals, medical interventions, and specific treatments or conditions, and covers many of the topics identified. It includes discussion surrounding healthcare staff such as nurses and doctors, as well as procedures such as surgeries and ultrasounds, and vast discussion around condition-specific care, including heart disease, cancer treatment, broken bones, pregnancy and birth, and hip replacement. This theme also covers patient health outcomes, through terms including \textit{healed} and \textit{improved}, as well as discussing post-treatment care, focussing on recovery processes, rehabilitation activities, and the role of exercise and therapy in patient recuperation through words such as \textit{rehab}, \textit{exercise}, \textit{speech}, \textit{physiotherapy}, \textit{balance}, \textit{medication}, \textit{discharged}, \textit{died}, \textit{passed}, and \textit{remission}. For example, a topic that relates to this general theme is a \textit{cancer} topic, using words such as \textit{cancer}, \textit{chemotherapy}, \textit{radiotherapy}, \textit{oncologist}, \textit{biopsy}, \textit{mass}, \textit{metastatic}, and \textit{remission}.

\textbf{Patient Experience, Emotion, Engagement, and Support}: This topic theme captures apparent emotional responses and comfort levels of patients, including feelings of anxiety, appreciation, and the support received from healthcare staff. It also describes the quality of the patient-caregiver interaction through discussion on personalised care, using both positive and negative words such as \textit{cared} and \textit{abused}. Communication is discussed, for example with terms such as \textit{listened}, \textit{told}, and \textit{rude}, as well as the emotional impact of the healthcare journey, through words such as \textit{safe}, \textit{comfortable}, and \textit{scared}. Patient support includes discussion on family members and personal relationships in the healthcare experience, using words such as \textit{family}, \textit{loved}, and \textit{dignity}, highlighting the presence of the support network in patient experience.

\textbf{Healthcare Environment, Operations, and Administration}: Topics within this theme discuss aspects of the healthcare environment, as seen with words such as \textit{ward}, \textit{bed}, and \textit{clean}. Additionally, discussion surrounding healthcare operations and logistics is present, featuring discussion around clinic appointments, waiting times, service accessibility, and administrative processes. This is evidenced by topics making use of words such as \textit{appointment}, \textit{admitted}, \textit{follow-up}, \textit{waiting}, \textit{sitting}, \textit{late}, \textit{system}, \textit{park}, and \textit{entrance}.

The identification of these themes through our topic modelling approach provides a complementary understanding of the patient-reported experience to traditional techniques, reflecting a comprehensive patient journey that spans clinical interactions, emotional responses, and the operational environment of healthcare delivery. 

For each topic, we find the likelihood of the topic given an emotion as $p(T = k | E = e)$ (Equation \ref{eq:topic-emotion-density}) and the likelihood of a topic given \textit{any} positive (or negative emotion $p(T=k | E \in E_p))$ (Equation \ref{eq:topic-sentiment-density}) and define topic positivity in Equation \ref{eq:ratio} as the ratio of positive topic likelihood to negative topic likelihood. 

\begin{figure*}[h!]
    \centering
    \begin{subfigure}[b]{0.5\textwidth}
      \centering
      \includegraphics[width=1\textwidth]{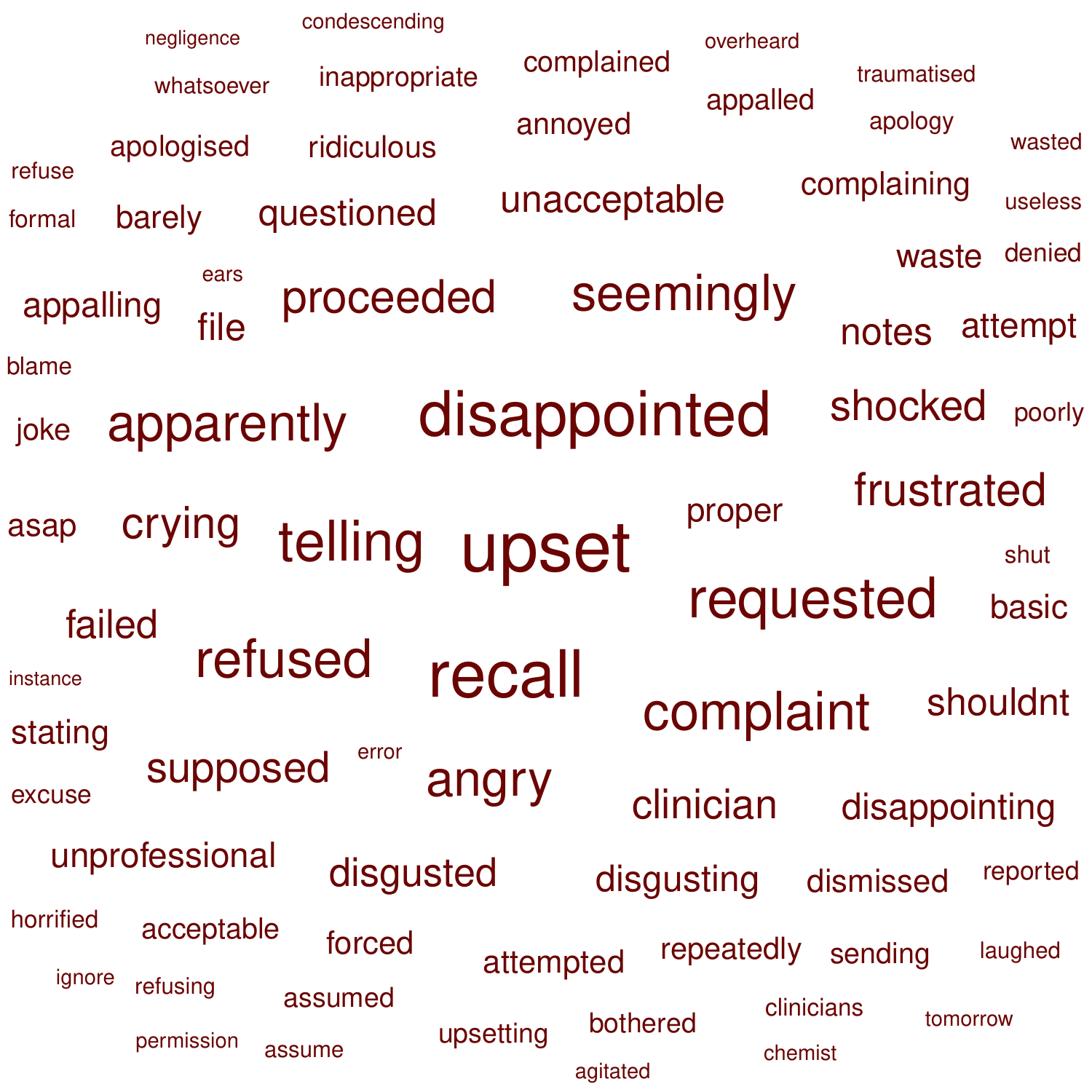}
      \caption{Topic with lowest positivity}
      \label{fig:upset}
    \end{subfigure}%
    \begin{subfigure}[b]{0.5\textwidth}
      \centering
      \includegraphics[width=1\textwidth]{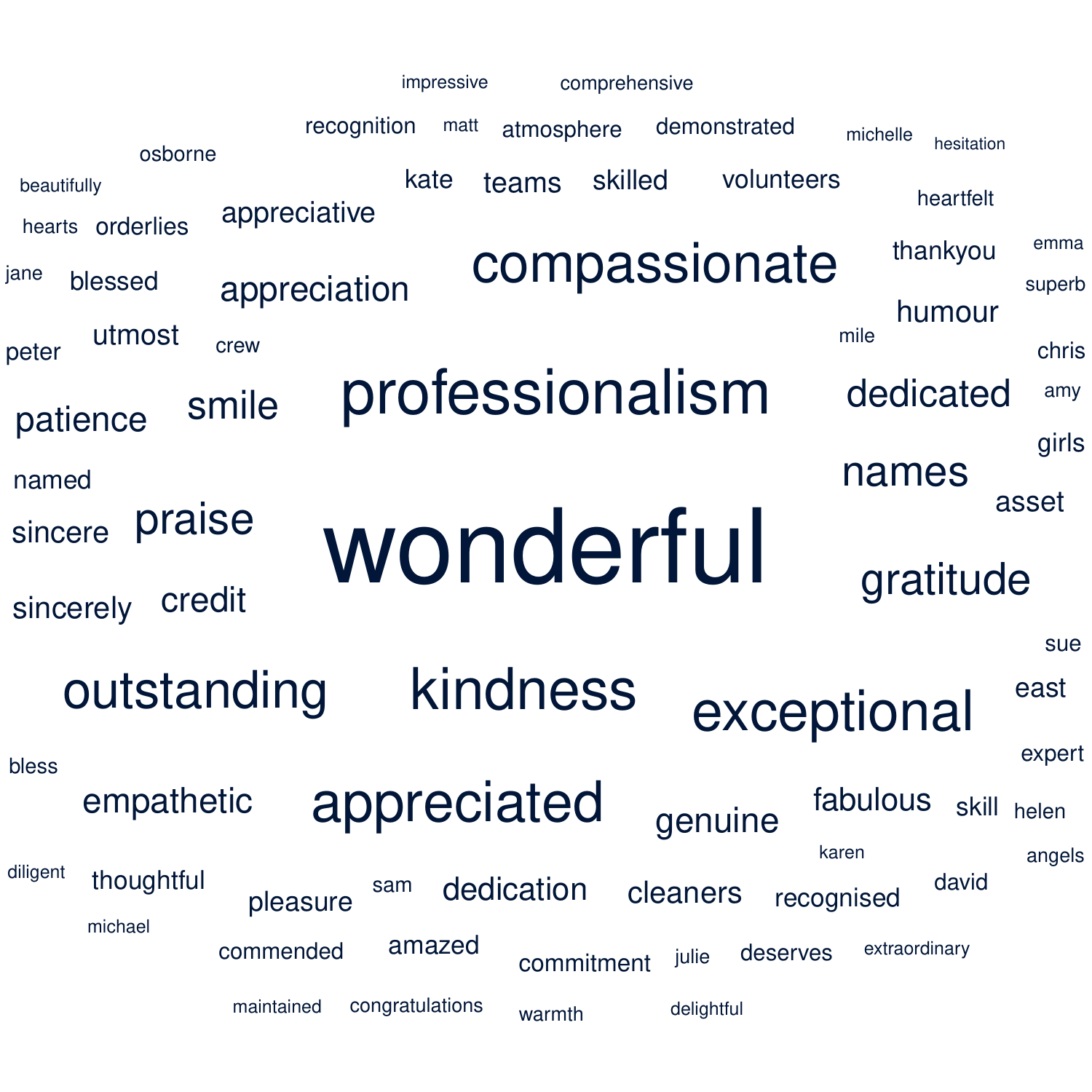}
      \caption{Topic with highest positivity}
      \label{fig:wonderful}
    \end{subfigure}
    \caption{Extremes in sentiment-topic associations in the Care Opinion corpus, illustrating the dichotomy of patient-reported experiences: (a) encapsulates critical negative experiences, while (b) reflects commendations. Both extremes centre around patient experience rather than patient outcomes.}
    \label{fig:extreme_positivity_ratios}
   \end{figure*}

We indicate the eight most positively-associated and eight most negatively-associated topics according to this ratio in Table \ref{tab:ratio}, along with their likelihoods $p(T = k | E\in E_p)$ (Positive) and $p(T = k | E\in E_n)$ (Negative). We visualise topics at the polar extremes of positivity in Figure \ref{fig:extreme_positivity_ratios}, with the lowest positivity in Figure \ref{fig:upset}, and the highest in Figure \ref{fig:wonderful}.

\begin{table*}[h!]
    \centering
    \caption{Topics and their conditional sentiment likelihoods for least-to-most positive topics from the Care Opinion corpus.}
    \label{tab:ratio}
    \footnotesize
    \begin{tabular}{lccr}
    \toprule
    Topic & \multicolumn{2}{c}{Likelihood} & \multicolumn{1}{c}{Positivity} \\
    \cmidrule(lr){2-3}
     & Positive & Negative & \\
    \midrule
upset, recall, disappointed, telling & 0.005 & 0.045 & 0.117\\
told, stated, replied, script & 0.001 & 0.009 & 0.156\\
front, rude, door, behaviour & 0.004 & 0.014 & 0.253\\
private, lack, request, form & 0.008 & 0.029 & 0.274\\
pay, cost, paid, afford & 0.001 & 0.005 & 0.283\\
waiting, wait, waited, sitting & 0.003 & 0.012 & 0.292\\
mri, chronic, spinal, spine & 0.001 & 0.004 & 0.305\\
booked, letter, date, list & 0.003 & 0.010 & 0.315\\
    \multicolumn{4}{c}{$\mathbf{\cdots}$} \\
support, helped, recovery, knowledge & 0.030 & 0.009 & 3.301\\
drug, lived, alcohol, addiction & 0.008 & 0.002 & 3.407\\
class, education, classes, online & 0.004 & 0.001 & 4.140\\
program, sessions, learnt, tools & 0.008 & 0.002 & 4.519\\
caring, professional, friendly, excellent & 0.036 & 0.006 & 5.842\\
amazing, grateful, comfortable, fantastic & 0.057 & 0.010 & 5.902\\
positive, supportive, recommend, supported & 0.016 & 0.003 & 6.207\\
wonderful, professionalism, kindness, appreciated & 0.052 & 0.008 & 6.388\\
\bottomrule
\end{tabular}
\end{table*}

Figure \ref{fig:extreme_positivity_ratios} reveals a dichotomy in patient-reported experiences within the healthcare system. The topic characterised by the lowest positivity, as depicted in Figure \ref{fig:upset}, uncovers a vast array of negative sentiments. The likelihood of this topic in the presence of negative emotions is quantified at 0.05, whereas it falls to 0.0045 for positive emotions, as detailed in Table \ref{tab:ratio}. It encapsulates a spectrum of adverse emotions such as \textit{upset}, \textit{disappointed}, and \textit{angry}, accompanied by descriptors such as \textit{unacceptable} and \textit{appalling}, and actions expressed through verbs such as \textit{questioned}, \textit{refused}, and \textit{forced}. Significantly, the emphasis of this topic is on the subjective experiences of patients rather than on the clinical outcomes of their care.

Conversely, the topic with the highest positivity (likelihood of 0.052 given a positive emotion compared to 0.008 for a negative emotion), illustrated in Figure \ref{fig:wonderful}, offers a contrasting portrayal. It accentuates the commendable attributes of healthcare workers, as evidenced by the prevalence of adjectives such as \textit{wonderful}, \textit{exceptional}, \textit{compassionate}, \textit{dedicated}, and \textit{empathetic}. Additionally, it includes nouns that convey a sense of appreciation, such as \textit{gratitude}, \textit{professionalism}, and \textit{kindness}, and is peppered with specific names, likely reflecting patient gratitude towards individual caregivers. 

A complete collection of the topics identified in our analysis --- along with their respective likelihoods under positive and negative sentiments --- is provided in the online supplementary materials, available on the dedicated GitHub repository \cite{Murray2024Supplement}.

\subsubsection{Relationships between Topics and Patient-Reported Emotions}

Interactions between patient-reported emotions and the topics patients discuss reveal the emotional impact that experiences in healthcare have on patients. Understanding the spectrum of emotions tied to themes in patient experience illuminates the subjective quality of healthcare encounters, informing providers and policy-makers on areas requiring compassionate engagement. Such insights contribute to developing patient-centred care models, enhancing communication strategies, and ultimately refining the landscape of healthcare delivery. 

    \begin{figure*}[!h]
        \centering
        \includegraphics[width=1\textwidth]{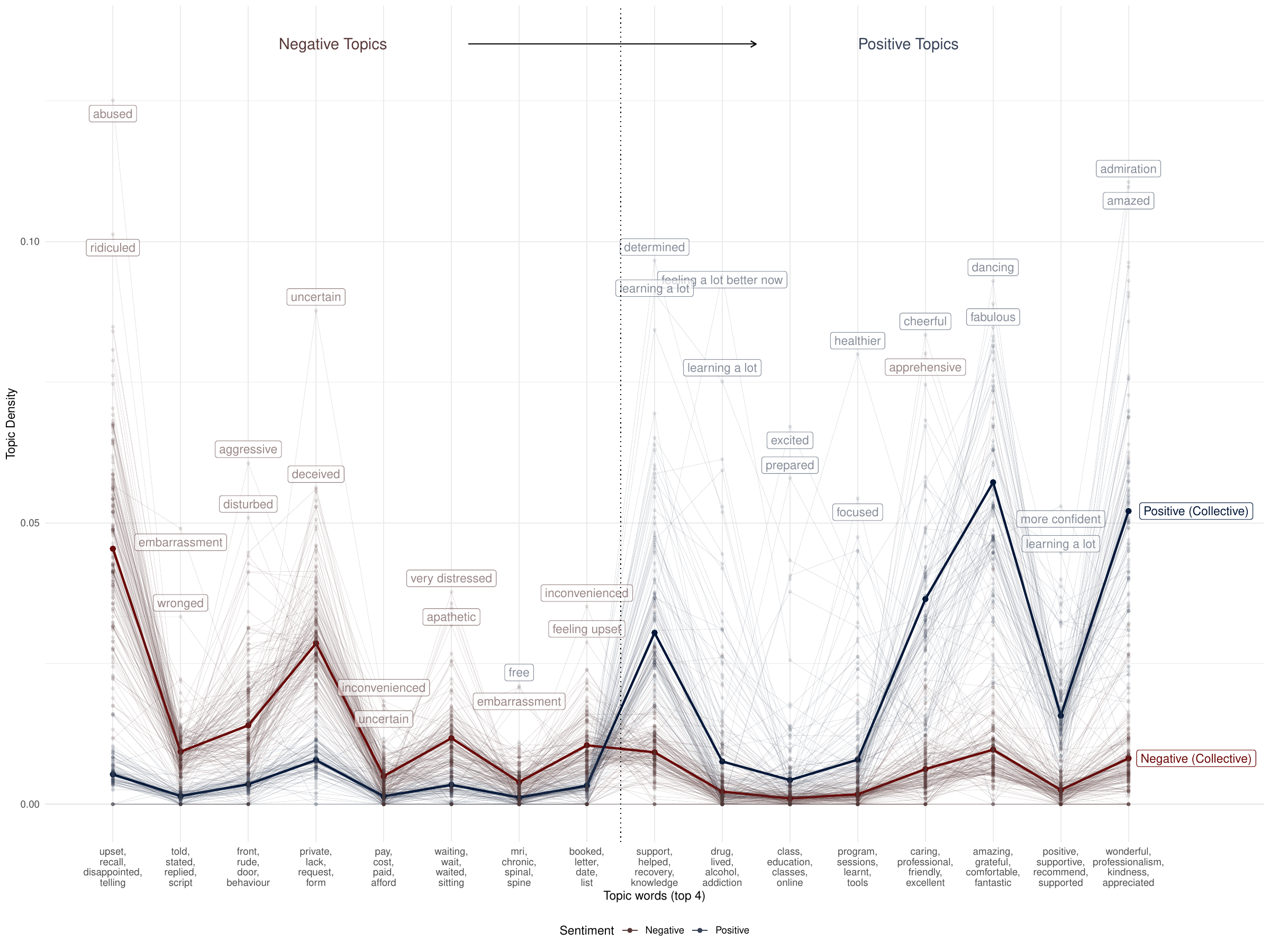}
        \caption{\textbf{Interactions between Emotions and Topics Discussed on Care Opinion}: This parallel coordinates plot illustrates the interplay between patient-reported emotions (depicted as lines) and Care Opinion topics (arranged on the x-axis), with the likelihood of each topic given the presence of specific emotions quantified on the y-axis. Among 105 total topics analysed, the plot selectively emphasises the eight most negative and eight most positive topics from left to right, highlighting the extremities of themes in patient narratives. The red spectrum signifies a collection of negative emotions, while the blue spectrum represents positive emotions. The strongest two emotions associated with each topic are labelled, providing clear indicators of the significant emotional drivers in patient experiences. The density and spread of lines corresponding to unlabelled emotions across the topics reveal the position and variation in emotional responses.}
         \label{fig:topic_density_sent}
       \end{figure*}

Figure \ref{fig:topic_density_sent} presents a parallel-coordinates plot that elucidates the relationships between themes in patient-reported experiences and their corresponding emotional responses. This visualisation captures the likelihood of specific topics conditioned on the spectrum of emotions expressed by patients, highlighting the prevalent sentiments tied to various facets of healthcare delivery. Along the x-axis, we list the eight most negatively-associated and eight most positively-associated topics from the patient reports. These are represented by their four most frequently used words, and arranged from left to right to correspond to an increase in topic \textit{positivity}, defined in Equation \ref{eq:ratio}. The y-axis of Figure \ref{fig:topic_density_sent} shows the topic-emotion likelihood $p(T=k | E = e)$ for each emotion $e$. The two emotions with the highest topic-emotion likelihood are labelled to reveal the emotions that are most strongly associated with each topic. The remainder of emotions are unlabelled, with a low opacity, and instead simply coloured to indicate positivity (blue) and negativity (red) to provide an overall view of the spectrum of positive and negative emotions for each topic. We add in bold the topic density marginalised over all positive emotions $p(T=k |E\in E_p)$, labelled \textit{Positive (Collective)} and all negative emotions $p(T=k |E\in E_n)$, labelled \textit{Negative (Collective)} to show the average topic-emotion likelihood for both sentiments.

\paragraph*{Negatively-Associated Topics:}

At the far left on the x-axis in Figure \ref{fig:topic_density_sent}, the topic represented by the terms \textit{upset}, \textit{recall}, \textit{disappointed}, and \textit{telling} -- previously identified in Figure \ref{fig:upset} -- carries the lowest positivity among all Care Opinion topics.  Predominantly, the emotions \textit{abused} and \textit{ridiculed} demonstrate the highest association with this topic, appearing to capture patients' extremely negative experiences and perceptions of care received from healthcare personnel. The overall negativity of this topic is evidenced by the distribution of negative emotions (red lines) sitting well above the distribution of positive emotions (blue lines). This pattern is evident for the following seven topics with low positivity. 

The pattern of negativity being most strongly exhibited in topics related to patient-experience, continues in the following two topics, with prominent emotions such as \textit{embarassment}, and \textit{aggressive} present. The fourth, fifth, sixth, and eighth topics (from left to right) appear to fall under the theme of \textit{Healthcare Environment, Operations, and Administration}, and elicit emotions such as \textit{uncertain}, \textit{inconvenienced}, and \textit{very distressed}. The emotion \textit{very distressed} is most strongly associated with a \textit{waiting} topic, indicating that high levels of distress are associated with patients having to spend time waiting for an aspect of care. The seventh topic, which mentions \textit{mri}, \textit{chronic}, and \textit{spinal}, as well as \textit{muscle}, \textit{neurology}, \textit{arthritis}, \textit{paralysed}, and \textit{chiropractor} is more closely related to the \textit{Clinical Care, Procedures, Recovery, Rehabilitation, and Outcomes} topic theme, and appears to discuss care around the musculoskeletal and nervous systems. Paradoxically, the positive term \textit{free} emerges as the most prevalent emotion within this context, a finding that may potentially reflect a sense of relief patients may experience upon recovery from enduring pain or mobility limitations. Conversely, the negative emotion \textit{embarrassed} also features prominently, potentially signalling the psychological distress or stigma that patients often confront when dealing with chronic illnesses. These dichotomous emotional responses underscore the complex interplay between the physical alleviation of symptoms and the social-emotional challenges encountered during the patient journey.

\paragraph*{Positively-Associated Topics:}

Most of the eight topics to the right of Figure \ref{fig:topic_density_sent} that exhibit the highest positivity of all topics, appear to fall under the theme of \textit{Patient Experience, Emotion, Engagement, and Support}. The rightmost, most positively associated topic (seen in Figure \ref{fig:wonderful}) is indicated by the words \textit{wonderful}, \textit{professionalism}, \textit{kindness}, and \textit{appreciated}. The likelihood of this topic is maximised for the emotions \textit{admiration}, and \textit{amazed}, showing signs of sincere patient gratitude. Several interesting observations can be made. The seventh most positive topic (from the right) appears to discuss addiction, which may fall more under the theme of \textit{Clinical Care, Procedures, Recovery, Rehabilitation, and Outcomes}. This topic, while appearing negative, may be a result of a shift from addiction to recovery. This is supported by the prominent emotions this topic exhibits, which are \textit{feeling a lot better now} and \textit{learning a lot}. The fifth and sixth most positive topics both are indicative of patient education and involvement, using words such as \textit{class}, \textit{education}, \textit{program}, \textit{sessions}, \textit{learnt}, and \textit{tools}. That this topic is of such high positivity indicates that patients who are actively included and educated in classes and programs express positive sentiments, with emotions such as \textit{excited}, \textit{prepared}, \textit{healthier}, and \textit{focused} appearing most prominently. The fourth most positive topic once again shows signs that positive patient-caretaker interactions, where the patient feels cared for, can have a beneficial response to patient concerns, as the negative term \textit{apprehensive} appears as one of the most prevalent emotions for this topic.

This analysis shows a clear delineation between topics associated with patient sentiment. Positive topics reflect quality interactions with healthcare workers and general experience, while those that are negatively connoted appear emblematic of adverse experiences within the healthcare system. In both cases, the extremes in positivity are strongly tied to patient-caregiver interactions rather than clinical outcomes. 

This detailed analysis is supported by comprehensive online supplementary materials, which present the full spectrum of topic-emotion relationships derived from the study. These materials are accessible for further review and exploration at the corresponding GitHub repository \cite{Murray2024Supplement}.

\subsubsection{A Landscape of Patient-Reported Emotions}

Our analysis through UMAP reveals two distinct clusters of emotions that show strong spatial separation and correspond to positive and negative emotions. In addition, we manually classify each emotion as either positive or negative and find that the UMAP clustering agrees with our manual labelling with an accuracy of 0.985.

\begin{figure*}[h]
 \centering
 \includegraphics[width=\textwidth]{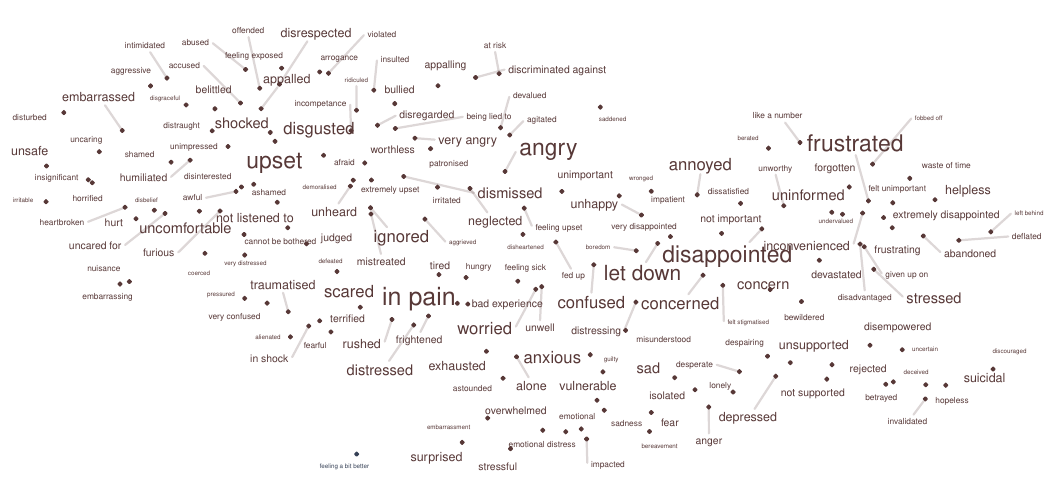}
 \caption{UMAP dimension reduction of emotion-topic representation (negative cluster).}
  \label{fig:umap_neg}
\end{figure*}

\begin{figure*}[!h]
    \centering
    \includegraphics[width=\textwidth]{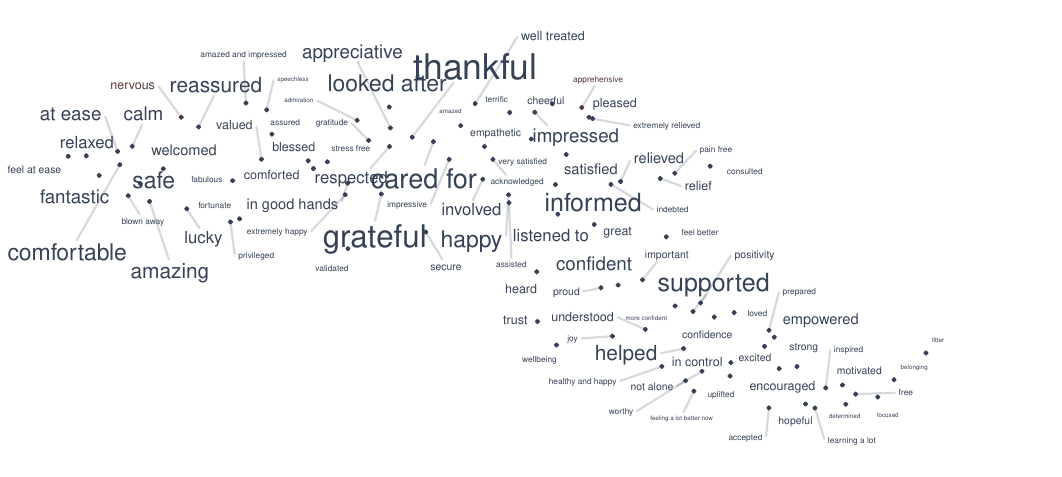}
    \caption{UMAP dimension reduction of emotion-topic representation (positive cluster).}
     \label{fig:umap_pos}
   \end{figure*}

Figure \ref{fig:umap_neg} and Figure \ref{fig:umap_pos} show the division of the UMAP-projected regions containing each cluster. Each emotion label in this figure is scaled according to the frequency of occurrences in the corpus and coloured by the manual sentiment classification—blue for positive and red for negative. This bifurcation of the emotion-topic space in its two-dimensional representation effectively captures the dichotomy between positive and negative emotions.  The closeness of emotions in the UMAP space suggests that there are thematic similarities in the narratives that elicited these emotions. Two emotions being close together in the UMAP visualisation implies that the patient experiences leading to these emotions share common themes or topics. 

\paragraph*{Negative Emotions:} 
This visualisation reveals several insights into patient experience. In the negative emotion cluster of Figure \ref{fig:umap_neg}, the co-occurrence of the intrinsic emotions \textit{frustrated} (upper right) and \textit{stressed} with the extrinsic emotions \textit{uninformed}, \textit{forgotten}, and \textit{inconvenienced} highlights the similarities and shared experience between these emotions that are characterised by neglect or lack of information. The proximity between \textit{angry} (upper middle) and the extrinsic emotions \textit{dismissed}, \textit{patronised}, and \textit{being lied to} reveals negative experiences that may cause patients to feel anger. Similarly, \textit{in pain}, \textit{scared}, \textit{frightened}, and \textit{distressed} appearing together shows that experiences resulting in traumatic emotions share similar themes.

The clustering (lower right) of \textit{suicidal}, \textit{depressed}, and \textit{hopeless} in close proximity to \textit{invalidated}, \textit{rejected}, and \textit{unsupported} demonstrates a significant correlation between profound negative emotional states and experiences of neglect or dismissal in patient narratives. This alignment not only reaffirms existing understanding of the impact of emotional validation (or lack thereof) on patient mental health, but also highlights the criticality of empathetic and supportive communication in healthcare settings.

\paragraph*{Positive Emotions:}

In the positive emotion cluster of Figure \ref{fig:umap_pos}, \textit{thankful} (upper middle), \textit{grateful}, \textit{looked after}, \textit{cared for}, \textit{respected}, and \textit{involved} are clustered together. This grouping suggests that experiences, where patients are cared for, included, and respected, are thematically similar to those where patients feel appreciation and contentment. This reinforces that fostering an environment of respect, inclusion, and quality care in a patient-centred care approach has a strong association with patients' satisfaction with the care they receive.

The positive clustering of \textit{informed} (lower middle), \textit{listened to}, \textit{heard}, \textit{trust}, and \textit{confident} illustrates the relationship between experiences where patients are actively engaged and those that they feel trust and confidence in. This finding emphasises the need for strong communication skills in addition to clinical competency to bolster patient confidence and trust.

Similarly, the proximity of \textit{supported} (middle right), \textit{helped}, \textit{encouraged}, and \textit{empowered}  with \textit{loved}, \textit{prepared}, and \textit{hopeful} reflects a narrative where patients feel actively and emotionally supported, as well as involved in their healthcare journey. This clustering suggests a relationship between these experiences and those where patients feel empowered and have a positive outlook towards their future.

The group comprising \textit{safe}, \textit{calm}, \textit{at ease}, and \textit{reassured} highlights the significant emotional impact that a secure and supportive healthcare environment can have on patients. This clustering suggests that when patients feel safe and reassured, it may cultivate a positive and calming environment. Notably, the emotion \textit{nervous} is also used within this positive cluster, an apparent paradox given its conventional connotation; however, in this context, it is plausible for patients to be nervous and yet report positive experiences if subsequently reassured. Similarly, \textit{apprehensive} (upper middle), a decidedly negative emotion, appears in the positive cluster situated next to \textit{extremely relieved}. This may be indicative of a potential shift from apprehension to relief, in a positive emotional transition in response to compassionate care.


\subsection{Probabilistic Emotion Recommender System}\label{sec:online_model}

We have developed a probabilistic emotion recommender system, as detailed in Section \ref{section:model}, which employs a network-based topic modelling approach on the Care Opinion corpus. By identifying distinct topics, the system projects new text inputs into a multidimensional topic space, with the resulting topic densities serving as predictors within a Naive Bayes classification framework. An interactive version of this model is available online \cite{persShiny}, complemented by a package in the R statistical computing environment \cite{team2013r}, which is freely available \cite{persR}.

\subsubsection{Model Evaluation} \label{sec:multiclass}

The evaluation of our probabilistic emotion recommender system is divided into two distinct categories. In this section, we assess the system's ability to predict specific emotions from the comprehensive range of patient-reported emotions, as derived in Equation \ref{eq:topic-emotion-density}. Later, we evaluate its performance in the binary classification task of discerning between positive or negative emotions, outlined in Equation \ref{eq:binary}.

Results from the $10$-fold cross-validation of the three probabilistic recommender systems, using metadata topic modelling, topic modelling, and the full-vocabulary model are presented, with comparisons to the baseline models using maximum likelihood estimates and uniform random guessing.

\begin{figure*}[h]
    \centering
    \begin{subfigure}[b]{0.5\textwidth}
      \centering
      \includegraphics[width=1\textwidth]{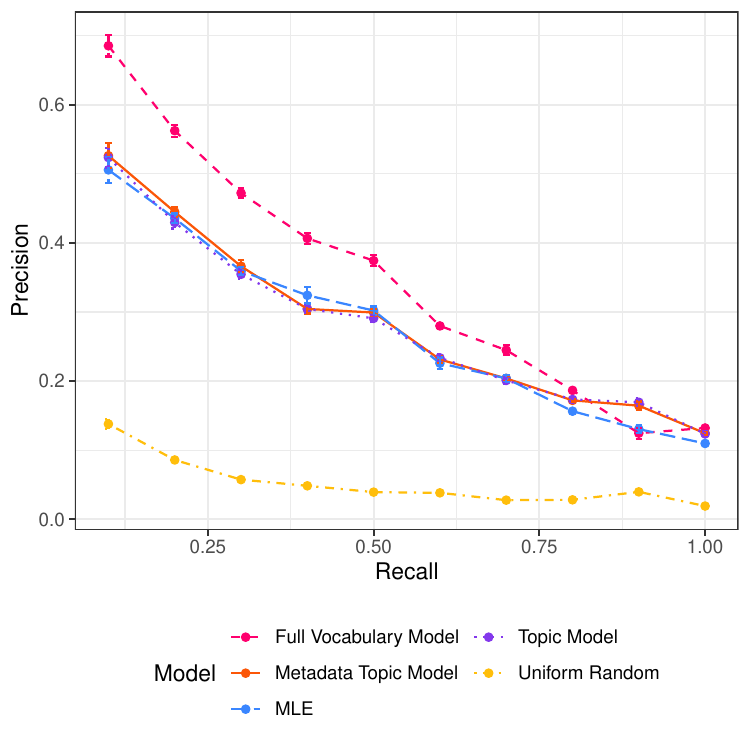}      \caption{Interpolated precision $P(r)$ at recall $r$.}
      \label{fig:prec_recall}
    \end{subfigure}%
    \begin{subfigure}[b]{0.5\textwidth}
      \centering
      \includegraphics[width=1\textwidth]{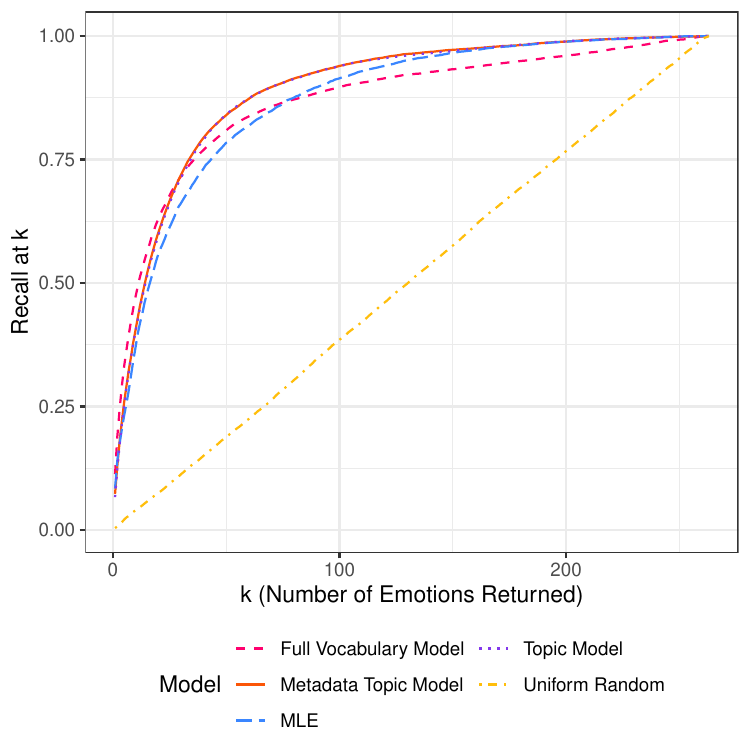}      \caption{Recall at $k$ for the $k$ highest posterior density emotions.}
      \label{fig:r_at_k}
    \end{subfigure}
    \caption{Evaluation of emotion recommender systems against baseline models for the Care Opinion corpus using precision and recall: The precision-recall plot in (a) shows strong performance of the full-vocabulary model, however, the recall at $k$ plot in (b) shows that when more than 30 emotions are returned, the models using dimension reductions are superior under this metric, potentially signalling overfitting in the full vocabulary model.}
    \label{fig:prec_rec_r_at_k}
   \end{figure*}

\paragraph{Precision and Recall:}
We find the interpolated precision \cite{manning2008introduction, zuva2012evaluation} $P(r)$ at recall $r$ as the highest precision for any recall $r^{\prime}$ greater than $r$,

$$P_i(r_i) = \max_{r_i \leq \prime{r_i^{\prime}}} p({r_i^{\prime}}),$$
and find the macro average across the validation sets. This is seen in Figure \ref{fig:prec_recall}, where we show standard errors of the macro averages across the folds with error bars. In Figure \ref{fig:r_at_k} we show the recall at $k$ (macro average across folds), where $k$ denotes the $k$ highest ranking emotions returned by the recommender system.

For small values of $k$ the full vocabulary model has the greatest recall, however, is outperformed by the topic model and metadata topic model when over $30$ emotions are returned. The full vocabulary model performs worse than simply selecting the highest density sentiment classes (MLE model) after $80$ emotions are returned. We argue that the higher performance in the full vocabulary model for small values of $k$ is due to the high sensitivity to detect specific emotions from specific mentions of words (e.g. if the word \textit{angry} appears it may conclude with near certainty that the text is of emotion \textit{angry}). 

The highly subjective nature of emotions in individuals suggests that model predictions with near certainty are likely a sign of overfitting. Predictions involving synonymous emotions, such as \textit{scared} and \textit{frightened} should account for this subjectivity with some density in prediction going to both emotions. However, potential overfitting in the full vocabulary model due to its high complexity may lead to predictions of single emotions with near certainty as a result of noise in the training data, amplified by the already sparse interactions between individual words and emotions. Poor model understanding of relationships between like-emotions may be missing when compared to more parsimonious approaches which meaningfully reduce the feature space, such as topic modelling.

\paragraph{Accounting for Partial Relevance and Penalising Late Arrival of Predicted Emotions:}

\begin{figure*}[!h]
    \centering
    \begin{subfigure}[b]{0.5\textwidth}
      \centering
      \includegraphics[width=1\textwidth]{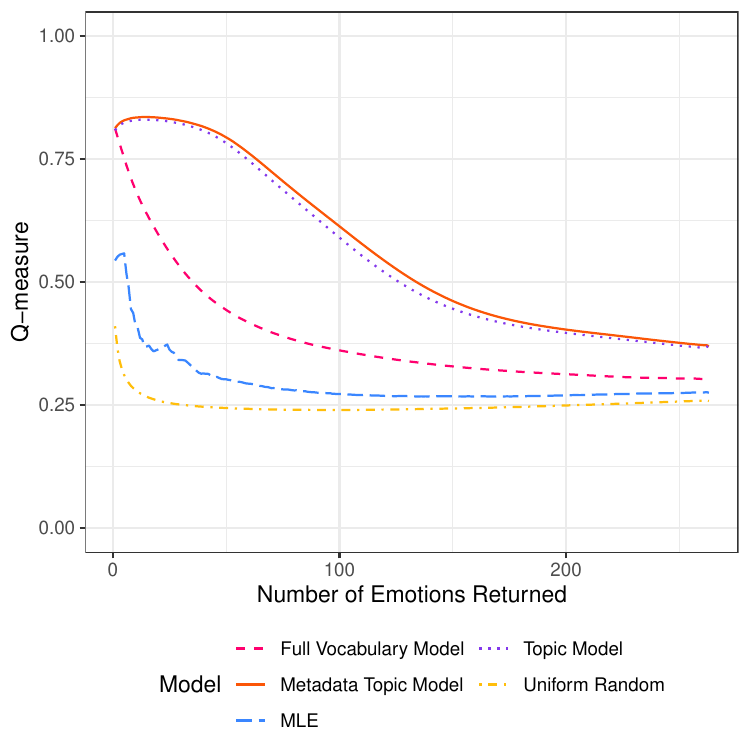}
      \caption{Q-measure for varying numbers of returned emotions.}
      \label{fig:q-measure}
    \end{subfigure}%
    \begin{subfigure}[b]{0.5\textwidth}
      \centering
      \includegraphics[width=1\textwidth]{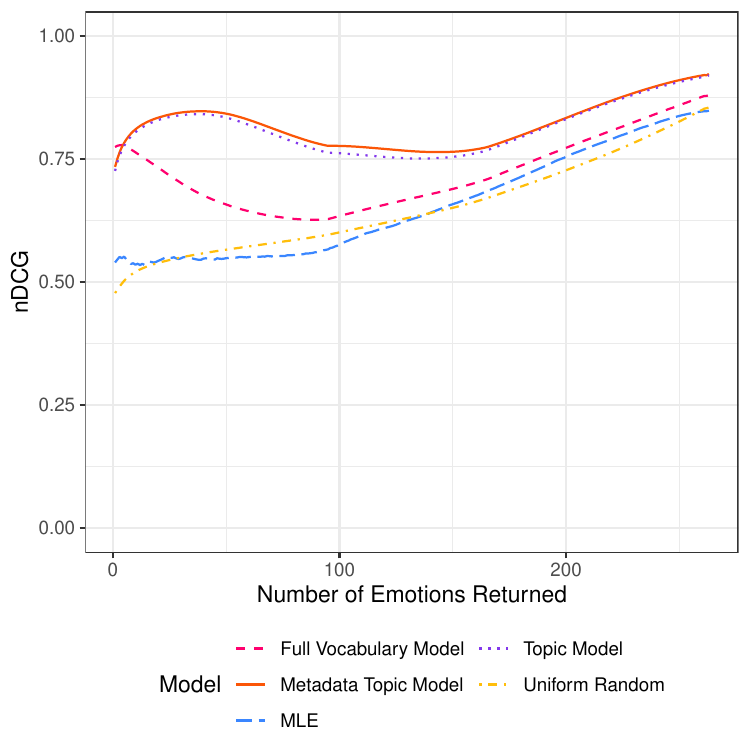}
      \caption{nDCG over the range of emotions returned.}
      \label{fig:ndcg}
    \end{subfigure}
    \caption{Comparative performance of emotion recommender systems with respect to the Q-measure (a) and nDCG (b). Each evaluation metric assesses the ranking order of emotions, with a score of 1 achieved under an ideal ranking of the preceding emotions. The metadata topic modelling shows the greatest performance under both metrics, whereas the high-complexity full-vocabulary model performs comparatively poorly.}
    \label{fig:q_and_ndcg}
\end{figure*}

We more appropriately validate these models by capturing partial relevance, as well as demonstrate greater performance in the topic modelling approaches by considering the degree of closeness between the predicted emotions and the labelled emotions in evaluation metrics. Precision and recall metrics fail here, as they will only recognise an emotion prediction as relevant if it is an exact match for the emotion label. This is a concern, as misclassification of \textit{frightened} when the labelled emotion is \textit{scared} is resultantly considered just as bad as predicting \textit{happy}. By considering partial relevance in model evaluation we can attribute value to predictions that have good, yet imperfect predictions. Graded relevance metrics, such as Q-measure and Normalised Discounted Cumulative Gain (nDCG) are more appropriate in evaluating information retrieval systems as they account for this partial relevance and moreover penalise the late arrival of relevant documents \cite{sakaiReliabilityInformationRetrieval2007}. By incorporating partial relevance in our evaluation metrics, as defined in Equation \ref{eq:rel}, we reduce the penalisation of predictions that are partially aligned with the true labels. Full definitions of Q-measure and nDCG may be found in \ref{app:eval} . Q-measure results for the probabilistic emotion recommender systems are shown in Figure \ref{fig:q-measure}, and nDCG results in Figure \ref{fig:ndcg}.

The data presented in Figure \ref{fig:q_and_ndcg} predominantly demonstrate the enhanced effectiveness of the metadata topic modelling approach, as evidenced by its performance in both Q-measure and nDCG metrics across the majority of the range for returned emotions. Notable exceptions are present: for instance, when only a single emotion is considered, both the topic modelling and full vocabulary approaches yield comparable Q-measure results; similarly, the full vocabulary model shows a slight advantage in nDCG when fewer than four results are returned. Nonetheless, these instances do not diminish the overall trend, which suggests that the dimension reduction inherent in topic modelling significantly bolsters the emotion recommender system's ability to discern associations among similar emotions. In contrast, the full vocabulary model exhibits a tendency toward overfitting, with its performance waning more sharply as the number of emotions increases, underscoring the robustness of dimension-reduced models in handling a broader spectrum of emotions. 

Additionally, both the topic models and the full vocabulary model demonstrate significant advancements over more rudimentary approaches, such as random guessing or defaulting to the most frequent emotions. This illustrates that, despite the complexity inherent in high-dimensional emotional data, the application of comparative metrics like Q-measure and nDCG can effectively showcase the relative performance enhancements afforded by these sophisticated models. Consequently, these models are validated as performing well beyond baseline levels, suggesting their practical utility in real-world applications.

\subsubsection{Binary Sentiment Classification of Patient Reports}

\begin{table*}
    \centering
    \caption{Evaluation metrics of the probabilistic emotion recommender system in the context of document-sentiment classification for three models compared to standard sentiment analysis lexicons. We show accuracy, balanced accuracy, F1, Precision, and Recall.}
    \label{tab:binary_eval}
    \footnotesize
\begin{tabular}{lccccl} 
\toprule
Model & Accuracy & Balanced accuracy & F1 & Precision & Recall\\
\midrule
Metadata Topic Model & 0.910 & 0.911 & 0.921 & 0.939 & 0.903\\
Topic Model & 0.908 & 0.911 & 0.918 & 0.945 & 0.893\\
Full Vocabulary Model & 0.877 & 0.869 & 0.896 & 0.878 & 0.914\\
Bing \cite{hu2004mining} & 0.846 & 0.844 & 0.812 & 0.794 & 0.830  \\
VADER \cite{hutto2014vader} & 0.842 & 0.822 & 0.785 & 0.845 & 0.727 \\
NRC \cite{mohammad2013crowdsourcing} & 0.705 & 0.647 & 0.491 & 0.788 & 0.357 \\
SentiWordNet \cite{baccianella-etal-2010-sentiwordnet} & 0.353 & 0.364 & 0.341 & 0.287 & 0.419 \\
AFINN \cite{NielsenF2011New} & 0.149 & 0.163 & 0.177 & 0.144 & 0.230 \\
\bottomrule
\end{tabular}
\end{table*}

Model evaluations of predictions made about binary sentiment using $P(E\in E_p | \mathbf{w})$ (and $P(E\in E_n | \mathbf{w})$) further capture the superior performance of the topic modelling approaches, and allow for evaluation using standard binary classification metrics. Evaluation metrics for sentiment classification of each model are shown in Table \ref{tab:binary_eval}, including metrics such as accuracy and balanced accuracy, as well as the macro averages across cross-validation folds of the F1 score, precision, and recall. Notably, the metadata-based topic modeling approach achieves the highest F1 score of 0.921. Among standard sentiment analysis lexicons, the Bing Sentiment Lexicon shows best performance, with an F1 score of 0.812, closely followed by VADER with 0.785. In contrast, models like SentiWordNet and AFINN exhibit significantly lower performance. This comparison illustrates the advantage of employing context-specific models over standard sentiment lexicons.

By framing the problem within a binary classification setting, we provide a complementary perspective to high-dimensional emotion profiling. This binary viewpoint allows for the utilisation of well-established metrics such as accuracy, precision, recall, and the F1 score, which offer a more intuitive and explainable measure of performance. In doing so, we enhance the interpretability of the results, facilitating a clearer understanding of the model's efficacy in distinguishing between positive and negative sentiments. Moreover, this approach demonstrates the versatility of the topic modelling techniques, which not only outperform the other models at high-dimensional emotion prediction but also provide tangible, interpretable outcomes when reduced to a binary sentiment classification framework. This dual analysis, spanning both high-dimensional and binary classification landscapes, provides a holistic evaluation of the models, enabling us to present a robust set of results that demonstrate the models' capabilities in both complex and simplified settings.

\subsubsection{Model Application to Simulated Patient-Reports}

In conjunction with rigorous model validation, we provide a supplementary demonstration of the probabilistic emotion recommender system's capabilities (using the metadata topic model) by applying it to simulated patient reports. The results, encapsulated in Table \ref{tab:sa}, showcase the input text and the top three emotions as determined by the model's posterior distribution, alongside empirical priors that align with the maximum likelihood estimates derived from the observed frequencies of emotions within the simulated reports. These results serve as an indicative snapshot of the model's potential to accurately identify emotions in textual data. We invite readers to further explore the model's utility via its interactive online version \cite{persShiny}, and R package \cite{persR} which accommodate the analysis of personalised text entries.

\begin{table*}[!h]
    \caption{Probabilistic modelling of emotions in simulated patient reports. This table displays the simulated report, the posterior of a positive sentiment, as well as the prior and posterior probabilities of the top three emotions identified by the model for each report.}
        \label{tab:sa}
    \centering
    \footnotesize
    \begin{tabular}{>{\raggedright\arraybackslash}p{25em}>{\raggedright\arraybackslash}p{5em}>{\raggedright\arraybackslash}p{5em}>{\centering\arraybackslash}p{4em}>{\centering\arraybackslash}p{4em}>{}p{4em}}
\toprule
Report & Positive Sentiment Posterior &  Emotion & Emotion Prior & Emotion Posterior\\
\midrule
 & & thankful & 0.071 & 0.194\\

 & & grateful & 0.050 & 0.155\\

\multirow[t]{-3}{25em}{\raggedright\arraybackslash Grandfather sadly passed away after a fall. Staff care was brilliant and family felt supported through a difficult time.} & \multirow[t]{-3}{4em}{\raggedright\arraybackslash 0.995} & supported & 0.032 & 0.139\\
\cmidrule{1-5}
 & & sad & 0.005 & 0.071\\

 & & upset & 0.021 & 0.057\\

\multirow[t]{-3}{25em}{\raggedright\arraybackslash Resident was left soiled and untreated for long periods of time while understaffed.} & \multirow[t]{-3}{4em}{\raggedright\arraybackslash 0.025} & angry & 0.022 & 0.046\\
\cmidrule{1-5}
& & hungry & 0.001 & 0.086\\

 & & disappointed & 0.024 & 0.065\\

\multirow[t]{-3}{25em}{\raggedright\arraybackslash The food was bland, cold, and made me feel sick.} & \multirow[t]{-3}{4em}{\raggedright\arraybackslash 0.117} & angry & 0.022 & 0.043\\
\cmidrule{1-5}
 & & inconvenienced & 0.002 & 0.215\\

 & & frustrated & 0.025 & 0.180\\
\multirow[t]{-3}{25em}{\raggedright\arraybackslash The patient experienced a long wait at a cancer clinic that resulted in doctors identifying a problem. The appointment was rescheduled without the patient knowing, and they felt as if they were being treated as the problem.} & \multirow[t]{-3}{4em}{\raggedright\arraybackslash 0.021} & annoyed & 0.009 & 0.110\\ 
& & & & \\
& & & & \\

\cmidrule{1-5}

 & & thankful & 0.071 & 0.128\\

 & & grateful & 0.050 & 0.107\\
\multirow[t]{-3}{25em}{\raggedright\arraybackslash The patient was scared by the prospect of surgery, however, felt that the staff's help put them at ease. They express their gratitude towards a particular practitioner for providing a high level of information.} & \multirow[t]{-3}{4em}{\raggedright\arraybackslash 0.986} & comfortable & 0.026 & 0.077\\
& & & & \\
\bottomrule
\end{tabular}
\end{table*}

The results from Table \ref{tab:sa} vividly illustrate the model's capacity to discern and quantify the emotions present within patient narratives. The model captures a high degree of positive sentiment in scenarios traditionally associated with negative emotions, such as the passing of a family member, where it identifies a strong sense of thankfulness and support despite the grief. Conversely, in situations typically expected to elicit strong negative responses, such as inadequate care or long wait times, the model not only confirms these expectations with low positive sentiment posteriors but also nuances the specific emotional responses, from sadness to frustration. This granular insight into patient emotions -- where a meal's poor quality evokes a notable sense of disappointment, and the fear of surgery is significantly alleviated by staff support -— highlights the model's potential to inform healthcare providers about the multifaceted emotional impacts of patient experiences, guiding more empathetic and responsive care practices.

This model has the potential to augment traditional patient-reported experience collection in healthcare by automatically labelling free-text comments in surveys. This adds distinct value to the harnessing of patient-reported experiences. As this model is context-specific to patient-reported experiences and demonstrates sound performance at both emotion recommendation and binary sentiment classification, the model facilitates a swift and accurate assessment of feedback, empowering healthcare providers to promptly address patient needs and sentiments. Beyond operational benefits, the model serves as a tool for embedding the patient voice more deeply into care evaluation and decision-making processes. Its high accuracy in both high-dimensional emotion recommendation and binary sentiment classification demonstrates its robustness and applicability. This dual capability ensures that both complex emotional profiles, as well as binary sentiments, are captured and quantified, providing a complementary understanding of the patient experience from the perspective of patient emotions. Looking ahead, the integration of this model into existing healthcare systems, supported by its accessibility through an interactive online platform \cite{persShiny} and an R package \cite{persR}, stands to significantly amplify the actionable insights derived from patient feedback.

\section{Discussion}

In this study, we use a dual-faceted approach to understanding and utilising patient-reported experiences in healthcare. Firstly, we undertake a detailed analysis of patient narratives sourced from Care Opinion. Secondly, we address the research problem by enhancing structured patient-reported experience mining with a scalable and transparent model. Our approach, which prioritises model interpretability, is designed to foster trust and encourage successful adoption in the analysis of patient-reported experiences, particularly in free-text comment analysis within patient-experience surveys.

\subsection{Analysis of Patient Narratives on Care Opinion}

Our findings indicate that while patient posts are predominantly positive, expressions of negative emotions are more intense and frequent. This suggests a complex emotional landscape where negative experiences in healthcare evoke more expressive and emotionally charged responses.

\subsubsection{Topic Themes in Care Opinion Reports}

Through topic modelling, we reveal relationships between themes in patient experiences and a spectrum of patient-reported emotions. The aggregation of topic densities for each emotion label provides a foundation for our subsequent analysis. 

We observe that topics fall under the three general themes, broadly relating to (1) \textit{Clinical Care, Procedures, Recovery, Rehabilitation, and Outcomes}, (2) \textit{Patient Experience, Emotion, Engagement, and Support}, and (3) \textit{
Healthcare Environment, Operations, and Administration}.

Collectively, these topics and their themes present a rich, patient-centred view of healthcare that goes beyond clinical outcomes to include the emotional and operational aspects of care. Such a holistic understanding of patient experience is informative to healthcare providers and policymakers as they strive to improve the quality of care and patient satisfaction. Our study contributes to this goal by providing a data-driven, patient-informed analysis that can inform future healthcare improvements and research directions.

\subsubsection{Relationships between Topics and Patient-Reported Emotions}

Additionally, our analysis reveals thematic differences in patient narratives associated with patient sentiment. Topics linked to positive sentiments reflect positive interactions with healthcare workers and general experience. Conversely, topics associated with negative sentiment typically reflect adverse experiences within the healthcare system. Notably, the most pronounced positive sentiments are frequently connected to the dynamics of patient-caregiver relationships, surpassing the influence of clinical outcomes.

Negative aspects of \textit{Healthcare Environment, Operations, and Administration}, such as \textit{cost}, and \textit{waiting} demonstrate a strong association with adverse emotional responses. In contrast, positive aspects within the same theme, notably initiatives aimed at enhancing patient engagement and education -- exemplified by educational programs -- elicit strong positive emotional reactions. Clinical care elements, particularly chronic and mobility-limiting conditions, are often associated with negative sentiments. Interestingly, discussions related to addiction recovery exhibit a trend towards positivity. This observation warrants cautious interpretation due to potential selection biases, as individuals in recovery might be more inclined to share their experiences on platforms such as Care Opinion.

These findings demonstrate the importance of interpersonal elements in shaping patient perceptions of care quality. They suggest that alongside clinical outcomes, the humanistic dimensions of healthcare delivery significantly influence patient satisfaction. Healthcare providers and researchers can leverage this data to pinpoint specific aspects of care that evoke strong emotional responses. This insight enables the development of targeted interventions to bolster patient satisfaction and emotional well-being.

\subsubsection{A Landscape of Patient-Reported Emotions}

By conducting a dimension reduction of aggregate topic profiles for each emotion, we revealed the dichotomy between positive and negative emotions in an emotional landscape. Our results suggest that profound negative emotional states like \textit{suicidal}, \textit{depressed}, and \textit{hopeless} are closely associated with experiences of neglect and dismissal, reinforcing the impact that emotional validation and supportive communication have on patient mental health. 

The convergence of such severe emotional states with feelings of invalidation and rejection suggests that these external experiences can be significant contributors to, or exacerbations of, deep-seated emotional distress. This finding emphasises the need for healthcare providers to be acutely aware of the power of their interactions with patients. The data indicates that negative experiences, such as feeling unsupported or invalidated, can have far-reaching consequences, potentially culminating in extreme emotional responses like feelings of hopelessness or suicidal ideation.

This insight has practical implications for patient care strategies, suggesting that interventions aimed at fostering a sense of validation, support, and acceptance could be crucial in mitigating negative emotional outcomes. It also underscores the importance of training healthcare professionals in emotional intelligence and communication skills, equipping them to recognise and respond to signs of such distress effectively.

On the positive side, the clustering of emotions such as \textit{thankful}, \textit{grateful}, and \textit{respected} with \textit{looked after} and \textit{involved} reinforces the beneficial impact of respect, inclusion, and comprehensive care on patient satisfaction. These insights suggest that enhancing communication and emotional support in healthcare settings can substantially influence patient well-being and perceptions of care quality. 

Additionally, the presence of typically negative emotion terms such as \textit{nervous} or \textit{apprehensive} within positive clusters may indicate a transition to positive states like relief, suggesting that the initial negative sentiment can be effectively addressed within a supportive healthcare environment. These findings offer a compelling argument for healthcare policies that prioritise emotional support and active patient engagement to foster positive patient experiences.

\subsection{Probabistic Emotion Recommender System}

This paper presents a novel contribution to the existing literature by developing and implementing a probabilistic emotion recommender system that functions in the context of patient-reported experiences. As this model is context-specific, it has a stronger justification for use in health care than non-contextual models that may be too generalised to offer specific health-related insights. Summarisation of documents into themes using topic modelling, that function as predictors for a probabilistic, Naive Bayes model affords model interpretability and transparency, both critical for healthcare applications. 

By considering the thematic overview that topic modelling affords, we wash away extraneous details that contribute to overfitting in a less-parsimonious model, arriving at a more generalisable model. We quantitatively show that the topic modelling dimension reduction affords superior performance in both emotion recommendation and binary sentiment classification when compared to a full-vocabulary model, through evaluation metrics under 10-fold cross-validation. These evaluation metrics, including Q-measure, and normalised discounted cumulative gain have strong information-theoretic justifications that allow for partial relevance and penalised late arrival of relevant results. Qualitatively, we have justified the topic modelling dimension reduction's interpretability by revealing relationships between the key themes of these documents and emotions.

In healthcare, the systematic evaluation of patient-reported experiences is critical for advancing quality improvement and fostering patient-centred care. The probabilistic emotion recommender system we have introduced represents a novel method for interpreting patient-reported experiences, offering a more detailed emotional analysis than what is typically achieved with binary sentiment analysis. This approach aims to provide healthcare providers with a more refined understanding of patient feedback, potentially aiding in the timely and targeted response to patient needs.

We readily anticipate the application of this model to augment traditional free-text patient-reported experience surveys, and online repositories, to provide immediate, accurate, and contextually relevant summaries of experience that reveal a spectrum of patient emotion. We enable this by providing our model as an R package (pers) \cite{persR} and online dashboard \cite{persShiny} that can easily be adopted by healthcare researchers and practitioners.

\subsection{Limitations}

This study, while offering rich insights into patient-reported experience and patient-emotion modelling is subject to several limitations that warrant consideration.

The reliance on patient narratives from the Australian online platform Care Opinion introduces demographic biases, potentially skewing our corpus towards certain age groups, socioeconomic statuses, or geographic locations, thereby making it non-representative of the broader population. Consequently, results derived from this data may not accurately reflect the broader patient population's experiences. A significant aspect of selection bias in our study is the likelihood of individuals with more extreme sentiments—either highly positive or negative—being more motivated to share their experiences. This is particularly evident in the context of addiction recovery stories. Narratives on Care Opinion discussing addiction often show a strong association with positive sentiments. However, this observation may be influenced by survivorship bias, as individuals who are successfully managing their addiction might be more inclined to post their stories online, potentially overshadowing the experiences of those still struggling or with less positive outcomes, leading to overly optimistic conclusions about the recovery process.

Viewing patient experience as a collective phenomenon can obscure the unique experiences, perceptions, and needs of individual patients. Healthcare is inherently personal, and what might be a positive experience for one patient could be neutral or negative for another. By aggregating experiences, nuanced details that are crucial to understanding each patient's perspective may be lost.

Additionally, the study's quantitative approach, while robust in many respects, might not capture the full depth of qualitative patient experiences. The transformation of rich, narrative data into quantifiable metrics risks losing the subtlety and individuality of patient experiences, a critical aspect of understanding patient-reported outcomes.

Finally, the practical application of the proposed probabilistic emotion recommender system in real-world healthcare settings poses its own set of challenges. Integration with existing healthcare systems, user adoption, and the real-world impact on healthcare outcomes are aspects that remain to be thoroughly evaluated. Additionally, ethical and privacy considerations in handling sensitive patient data are paramount and require diligent attention to ensure responsible use of the technology.

\subsection{Future Research}

Future research could expand upon these findings, employing longitudinal data to track the evolution of patient experiences over time and through various healthcare reforms. For instance, tracking topic densities associated with known positive and negative emotions, as well as the prevalence of specific patient-reported emotions, or overall sentiment can supplement traditional healthcare performance monitoring. This could be especially effective in response to specific healthcare interventions or policy changes. Additionally, comparative studies between different healthcare systems or regions could offer insights into systemic influences on patient experiences. Further, implementing and evaluating the probabilistic emotion recommender system in real healthcare settings would provide practical insights into its effectiveness, user adoption, and impact on healthcare outcomes. Investigating how the probabilistic emotion recommender system can be used to tailor healthcare services to individual patients, such as through personalised interventions based on patient-reported emotions.

While our study sheds light on key aspects of patient care, we acknowledge its limitations, including potential biases inherent in self-reported data and the need for broader demographic representation. Further studies should aim to corroborate these insights with a more diverse patient cohort.

\section{Conclusion}

This study makes significant contributions to the evolving landscape of healthcare, where understanding and integrating patient experiences and emotions are becoming increasingly crucial for quality care. Our main contributions are (1) an analysis of the free-text narratives on the popular healthcare review website Care Opinion, and (2) the development and implementation of a probabilistic emotion recommender system. 

In our analysis of patient experiences on Care Opinion, we conduct topic modelling to summarise patient experience into topics, broadly relating to themes in clinical care, patient experience, and healthcare logistics. By capturing relationships between patient-reported emotions and their aggregate thematic composition, we reveal numerous insights into patient-reported experiences that can help inform healthcare practitioners and researchers. We show that topics linked with both extremely high and low degrees of positive sentiment are most closely related to areas in subjective patient experience, such as interactions with healthcare staff, rather than outcomes of clinical care. Additionally, we reveal a landscape of patient-reported emotions, that helps contextualise relationships between specific patient-reported emotions. For instance, we show that extremely negative self-reported emotion terms that capture intrinsic states, such as suicidality, depression, and hopelessness, exhibit a strong relationship with negative emotions relating to experiences such as dismissal, invalidation, and rejection. Our findings indicate that patient experience is greatly improved through positive patient-caregiver interactions and initiatives that increase patient engagement, such as educational programs. Additionally, we show that positive patient-caregiver interactions are associated with mitigating negative emotions such as apprehension. Our study also shows that strong states of negativity are present in experiences with chronic and mobility-limiting conditions, however, experiences in healthcare relating to addiction tend to be framed positively.

In addition to the patient narrative analysis, we develop a probabilistic emotion recommender system from the Care Opinion reports and their corresponding patient-reported emotion labels. This context-specific tool, leveraging topic modelling in a probabilistic Naive Bayes approach, demonstrates not only a high degree of interpretability and transparency, but also superior performance in emotion recommendation and binary sentiment classification compared to comparator and baseline models, including standard sentiment lexicons. This is seen through our quantitative evaluations, including 10-fold cross-validation under appropriate metrics such as Q-measure and normalised discounted cumulative gain. The system distillation of complex narratives into thematic composition both (a) reduces the risk of overfitting, enhancing its applicability and generalisability in real-world healthcare settings and (b) improves interpretability through the use of meaningful features.

Our model's proficiency in identifying and classifying emotions within patient narratives and interpretability positions it as a valuable asset for healthcare providers. This model facilitates more nuanced patient care and service improvement by providing a deeper understanding of how patient experiences link with emotions in a patient-centred care framework. By offering healthcare professionals a transparent, probablistic tool to comprehend patient feedback comprehensively, available as an R package and online dashboard, it opens avenues for more tailored and empathetic patient care strategies and allows for the augmentation of free-text comments on patient-reported experience surveys. This approach aligns with the current healthcare emphasis on patient-centeredness and helps to enhance patient-provider communication. Future research should focus on expanding the model's application to more diverse healthcare contexts and exploring its utility in longitudinal patient experience studies.

\section*{Ethical Approval}

Not applicable. As this study involves the analysis of publicly available data from Care Opinion (Australia), which is an open platform where patients share their healthcare experiences, which are classified as a public resource intended for broader public benefit by Care Opinion, this study is not considered human subjects research. Therefore, no consent for this research was required. In line with Care Opinion's terms of re-use under the Creative Commons licence Attribution-\-Non\-Commercial-\-ShareAlike 4.0, our research strictly adheres to non-commercial scholarly purposes, ensuring that individual privacy is respected, identifying information is not disclosed, and any patient narratives that we mention are entirely simulated.

\section*{Declaration of Competing Interests}
The authors declare that they have no known competing financial interests or personal relationships that could have influenced this study.
    
\section*{CRediT Author tatement}
\textbf{Curtis Murray:} Conceptualisation, Methodology, Software, Formal analysis, Data Curation, Writing - Original Draft, Writing - Review \& Editing, Visualization. \textbf{Lewis Mitchell:} Conceptualisation, Methodology, Supervision, Writing - Review \& Editing Funding acquisition. \textbf{Simon Tuke:} Conceptualisation, Methodology, Supervision Writing - Review \& Editing. \textbf{Mark Mackay:} Conceptualisation, Methodology, Supervision, Writing - Review \& Editing.
    
\section*{Funding}
The authors disclosed receipt of the following financial support for the research, authorship, and/or publication of this article: This work was supported by The University of Adelaide, LM is supported by the Australian Government through the Australian Research Council’s Discovery Projects funding scheme (project DP210103700).

\section*{Acknowledgements}

We would like to acknowledge the authors who shared their experiences on Care Opinion. In addition, we would like to thank the staff at Care Opinion for creating an open platform for patients to voice their experiences.

\appendix{}

\section{Probabilistic Emotion Reccomender System} \label{app:pers}

For the purposes of this appendix, when we refer to probabilities in shorthand, where the random variable is assumed. For example,  $p(A=a | B=b)$ is written succinctly as $p(a|b)$ and the random variables are inferred by context.

Probabilistic modelling of the emotional distribution of text can be conducted from the posterior distribution $E|d \sim p(e|d)$, where $e$ is an emotion and $d$ is a document in bag-of-words representation. This is equivalent to $E|d \sim p(e | \mathbf{w})$, where $\mathbf{w}$ is the bag-of-words vector, or $E|d \sim p(e | w_1, \cdots , w_{|V|})$, where $w_i$ is the word count of the word at index $i$ in vocabulary $V$. Using Bayes' Theorem, we rewrite the posterior distribution as
$$
p(e| \mathbf{w}) = \dfrac{p(\mathbf{w} |e)p(e)}{\sum_{j =1}^{n}p(\mathbf{w} | e_j)p(e_j)}, 
$$
where the priors $p(e_j)$ are taken as the maximum likelihood estimates in an empirical Bayesian approach \cite{collins2013naive}. 

Since $|V|$ is typically in the order of thousands or tens of thousands, the feature space is large, making density estimates have poor generalisability \cite{hastie2009elements}. A typical simplification that is often made in Natural Language Processing is an assumption of class-conditional independence of words in a naive Bayes approach \cite{mitchell1997machine}.

Under this assumption, the conditional density is 
\begin{align*}p(\mathbf{w} | e) &= p(w_1| w_2, \cdots, w_n, e) \cdots p(w_n | e)\\
&= p(w_1| e)\cdots p(w_n | e)\\
& = \prod_{ \{ i: w_i \neq 0 \} } p( w_i | e)
\end{align*}
So that
\begin{align}\label{eq:word}
p(e | \mathbf{w}) =  \dfrac{
\left( \prod_{\{ i: w_i \neq 0 \}} p( w_i | e) \right) p(e)}{\sum_{j =1}^{n} \left( \prod_{\{ i: w_i \neq 0 \}} p( w_i | e_j)\right) p(e_j)}
\end{align}
However, even with the reduction in the dimensionality of the feature space, the model is not particularly parsimonious, having $|V| \times |E|$ parameters, and may cause overfitting. Topic modelling can act as a dimension reduction tool, taking high dimensional documents, to relatively lower dimensional topic mixtures $TM: \mathbb{N}^{|V|} \to [0,1]^{n_t}$, $\mathbf{w} \mapsto \mathbf{t}$, where $\mathbf{t}$ is a $n_t$-dimensional vector of topic densities $p(t_k | \mathbf{w})$, and topics are $|V|$-dimensional word mixtures with elements $p(v_i | k)$ for topics $T$, with $v_i \in V$. 


Under the dimension reduction, we can estimate
\begin{align}\label{eqn:exponent}
p(w_i | e) = \sum_{k = 1}^{n_t}p(v_i | k, e)^{w_i}p(k| e)^{w_i},
\end{align}
where the topic-emotion distributions from Equation \ref{eq:topic-emotion-density}. Note the use of the exponent $w_i$ in Equation \ref{eqn:exponent}, the word count of word $v_i$, to account for words with word counts exceeding one. 

Topic models such as those found through network-based topic modelling allow topics to partition the vocabulary, i.e. words belong to exactly one topic \cite{gerlach2018network}. This allows the summation in (\ref{eqn:exponent}) to collapse to the term with non-zero probability. This term corresponds to the topic that word $w_i$ belongs to, say, $k_{w_i}$;
$$
p(w_i | e) = p(v_i | k_{w_i}, e)^{w_i}p(k_{w_i}| e)^{w_i}.
$$

Further, we make an assumption that words are conditionally independent to emotions given their topic, that is to say, that  
$$p(v_i | k_{w_i}, e) =  p(v_i | k_{w_i}).$$

This gives,
\begin{align*}
    & p(e | \mathbf{w}) = \\& \dfrac{\left( \prod_{\{ i: w_i \neq 0 \}} p(v_i | k_{w_i})^{w_i}p(k_{w_i}| e)^w_i \right) p(e)}{\sum_{j =1}^{n} \left( \prod_{\{ i: w_i \neq 0 \}} p(v_i | k_{w_i})^{w_i}p(k_{w_i}| e_j)^{w_i}\right) p(e_j)},
\end{align*}
which can be simplified by noticing that the term $$\prod_{\{ i: w_i \neq 0 \}} p(v_i | k_{w_i})^{w_i}$$ can be pulled out from the summation in the denominator as it does not depend on $j$, and hence cancels with the corresponding term in the numerator (it is always non-zero as $v_i$ is of topic $k_{w_i}$ by construction),

\begin{align}\label{eqn:numerator}
p(e | \mathbf{w}) =  \dfrac{\left( \prod_{\{ i: w_i \neq 0 \}} p(k_{w_i}| e)^{w_i}\right)p(e)}{\sum_{j =1}^{n} \left( \prod_{\{ i: w_i \neq 0 \}} p(k_{w_i}| e_j)^{w_i}\right) p(e_j)}.
\end{align}
Numerical underflow as a result of repeated multiplication of near-zero numbers has the potential to influence these results. In order to combat this, we employ the log-sum-exp
trick to avoid numerical underflow \cite{murphy2006naive}. This is illustrated below.

By taking the log of the numerator in Equation \ref{eqn:numerator} the log of the products can be exchanged for the sum of the logs;

\begin{align*}
    & \log\left(\left( \prod_{\{ i: w_i \neq 0 \}} p(k_{w_i}| e)^{w_i}\right)p(e)\right) \\
    = & \sum_{\{i: w_i \neq 0 \}} \log\left(w_i\;p(k_{w_i}| e)\right) + \log\left(p(e)\right)
\end{align*}

This exchange prevents the numerator from underflowing as the repeated product of near-zero probabilities is avoided. If we seek to find $\log\left(p(e | \mathbf{w})\right)$ by employing a similar strategy in the denominator,

\begin{equation}
\begin{aligned}
     & \log\left(p(e|\mathbf{w})\right) = \sum_{\{i: w_i \neq 0 \}} \log\left(w_i\;p(k_{w_i}| e) +  \log\left(p(e)\right)\right) - \\
&  \log  \sum_{j=1}^n \exp\left(\sum_{\{i: w_i \neq 0 \}} \log\left(w_i\;p(k_{w_i}| e_j)\right)+ \log(p(e_j))\right),
\end{aligned}
\end{equation}

we see that the later term, say $U$,  
\begin{align} \label{term:exp}
    & U = \nonumber\\
    & \log  \sum_{j=1}^n \exp\left(\sum_{\{i: w_i \neq 0 \}} \log\left(w_i\;p(k_{w_i}| e_j)\right)+ \log(p(e_j))\right) 
\end{align}
has the potential to underflow due to the exponentiation of large negative numbers that result from the summation of the log of many near-zero numbers. Fortunately, we can avoid this numerical underflow as follows. Firstly, let $S_j$ denote the terms inside the exponentials of $U$,
$$S_j = \sum_{\{i: w_i \neq 0 \}} \log\left(w_i\;p(k_{w_i}| e_j)\right)+ \log(p(e_j))$$
so that 
\begin{align} \label{term:exp_U}
U = \log \sum_{j=1}^n \exp(S_j),
\end{align}
where each of the $S_j$ is negative by definition. By subtracting the largest of these negative sums,
$$\hat{S} = \max_j S_j,$$
from within each of the exponentials of $U$ in Equation \ref{term:exp_U}, and accounting for this with a tactful multiplication of the corresponding exponential

\begin{align}
    U &= \log \sum_{j=1}^n \exp(S_j-\hat{S})\exp\left(\hat{S}\right)\nonumber\\
    &= \hat{S} + \log \sum_{j=1}^n \exp(S_j-\hat{S}),\label{eqn:wont_underflow}
\end{align}
 numerical underflow is avoided, as the most dominant term is captured directly without needing to exponentiate the log of a small number. In full, this gives the log posterior as

\begin{equation}
    \begin{aligned}
        &\log\left(p(e | \mathbf{w})\right) = \\
        &   \sum_{\{i: w_i \neq 0 \}} \log\left(w_i\;p(k_{w_i}| e)\right) + \log\left(p(e)\right) - \hat{S} - \\
& \log  \sum_{j=1}^n \exp\left(\sum_{\{i: w_i \neq 0 \}}  \log\left(w_i\;p(k_{w_i}| e_j)\right)+ \log(p(e_j))-\hat{S}\right).
    \end{aligned}
\end{equation}

Additionally, as there is the potential for only few documents being associated with a particular emotion $e$, it may arise that $p(k_{w_i}| e) = 0$ for some topics $k$, resulting in posterior $p(e|\mathbf{w})$ collapsing to zero whenever the word $v_i$ appears. We avoid this by adding a small, non-zero number to each $p(k_{w_i}| e) = 0$.

\subsection{Evaluation Metrics} 
\label{app:eval}

When considering documents labelled with multiple emotions, the gain for the $r^\text{th}$ ranked emotion, $g(r)$, is computed as follows:

$$
g(r) = \max_{e_l \in L} \mathrm{rel}(e_r, e_l),
$$

where $e_r$ is the ranked emotion, and the calculation seeks the highest relevance score between $e_r$ and the most closely related label emotion, $e_l$, in the given set of labels $L$.

\paragraph{Q-measure:}

Q-measure is a generalisation of average precision in a binary setting to account for partial relevance, defined as
\begin{align}\label{eq:qmeasure}
Q(k) = \frac{1}{k}\sum_{r = 1}^k \mathbbm{1}_{\mathrm{rel}(r) > 0}\text{Br}(r),
\end{align}

where $Br(r)$ is the \emph{blended ratio},
\begin{align}\label{eq:br}
    Br(r) = \frac{\text{cg}(r) + \sum_{i=1}^r\mathbbm{1}_{g(i) > 0}}{\text{cg}_I(r) + r},
\end{align}
$cg(r)$ is the \emph{cumulative gain} of the top $r$ ranked emotions,
\begin{align}\label{eq:cg}
    \text{cg}(r) = \sum_{i = 1}^r g(i),
\end{align}
and $\text{cg}_I(r)$ is the cumulative gain of the top $r$ ranked emotions in an ideal ranking, i.e. that which ranks results with non-increasing relevance for increasing rank. 

\paragraph{Normalised Discounted Cumulative Gain (nDCG)}

Another metric that considers partial relevance, as well as penalises the late arrival of relevant documents is nDCG. We first introduce discounted cumulative gain (DCG), which also allows for the penalisation of the late arrival of relevant documents by \emph{discounting} the $r^{\text{th}}$ ranked result's gain according to the rank. Traditionally, the discounting factor is logarithmic, and relevant retrieval is more strongly emphasised by exponentiating the gain,
$$
\text{DGC}_r = \sum_{i=1}^r\frac{2^{g(i)}-1}{\log_2(i+1)}.
$$
DCG should not be used to grade information retrieval systems across differing queries, as some queries may have more (or less) potential relevant results to select from, resulting in a greater (or lesser) DCG. To account for this, DCG is normalised by the \emph{ideal} discounted cumulative gain (IDCG), the maximum DCG achievable, resulting from an ideal document ranking with non-increasing relevance as rank increases,
$$
\text{nDCG}_r = \frac{\text{DCG}_r}{\text{IDCG}_r}.
$$

\section{Calculating the posterior distribution of emotions given a document}\label{appendix:example}

Here we illustrate how to compute the vector of emotion-densities $$\left[ p(E = e_j | \mathbf{w}) \right]_j$$ using the network topic model in Figure \ref{fig:model-example} where we use words from document $d_1$; $\mathbf{w_1} = (2,1,0,1)$.

First, we find the empirical densities of each emotion class. For each emotion, this is the number of edges out of the emotion, divided by the total number of edges out of all emotions. For example, there are two edges out of emotion $e_1$, out of the combined five edges out of all emotions, so $p(E=e_1) = \frac{2}{5}$. In full, the column vector $[p(E = e_j)]_j$ is

\begin{align}\label{eq:sent}
\left[p(E = e_j)\right] = \begin{bmatrix}\frac{2}{5}\\[0.3em]
    \frac{1}{5}\\[0.3em]
    \frac{2}{5}
    \end{bmatrix}.
\end{align}

The matrix $\left[p(T = t_i|d_j) \right]_{i,j}$ shows the empirical topic-class use for each document. For example, Document 1 uses topic $t_1$ three times (two occurrences of word $v_1$ and one of $v_2$), and topic $t_3$ once ($v_4$ is used once). This gives $p(T=t_1 | d_1) = \frac{3}{4}$. In full, we find
\[
\left[p(T = t_i|d_j) \right]_{i,j} = \begin{bmatrix}
    \frac{3}{4} & 1 & \frac{2}{3} & 0 \\[0.3em]
    0 & 0 & \frac{1}{3} & \frac{2}{3}\\[0.3em]
    \frac{1}{4} & 0 & 0 & \frac{1}{3}\\
\end{bmatrix}.
\]

The matrix $\left[p(d_i|e_j) \right]_{i,j}$ tells us the probabilities of selecting a random document $d_i$ belonging to an emotion $e_j$. Since each document is assumed equally likely, each document has a probability equal to the inverse of the emotion's use:
\[
\left[p(d_i |E = e_j) \right]_{i,j} = \begin{bmatrix}
    \frac{1}{2} & 1 & 0\\[0.3em]
    \frac{1}{2} & 0 & 0\\[0.3em]
    0 & 0 & \frac{1}{2}\\[0.3em]
    0 & 0 & \frac{1}{2}\\
\end{bmatrix}.
\]

This allows us to calculate $\left[p(t_i | e_j) \right]_{i,j}$:

\begin{align*}
\left[p(t_i | e_j) \right] &= \left[p(T = t_i|d_j) \right]_{i,j}  \left[p(d_i |E = e_j) \right]_{i,j} \\
&= \begin{bmatrix}
    \frac{7}{8} & \frac{3}{4} & \frac{1}{3}\\[0.3em]
    0 & 0 & \frac{1}{2}\\[0.3em]
    \frac{1}{8} & \frac{1}{4} & \frac{1}{6}\\
    \end{bmatrix}.
\end{align*}

Note that the implicit summation in the matrix multiplication here marginalises over documents. This likelihood, and the emotion-class densities in Equation \ref{eq:sent} can be used with the model in Equation \ref{eqn:numerator} to find the posterior distribution:
\begin{align*}
    \left[p(E=e_j | d_1)\right]_{j} \approx \begin{bmatrix}
    0.59\\
    0.37\\
    0.04
    \end{bmatrix}.
\end{align*}

\bibliographystyle{elsarticle-num} 
\bibliography{bibliography}

\end{document}